\documentclass[journal,comsoc]{IEEEtran}
\usepackage[T1]{fontenc}

\ifCLASSINFOpdf
\else
\fi
\usepackage{amsmath}
\usepackage[cmintegrals]{newtxmath}
\usepackage{hyperref}
\usepackage{url}

\usepackage{algorithm}
\usepackage{bm}
\usepackage{algorithmic}
\usepackage{graphicx}
\usepackage{textcomp}
\usepackage{subfigure}
\usepackage{makecell}
\usepackage{balance}
\usepackage{multirow}
\usepackage{booktabs} 
\usepackage{bbm}

\begin{document}
%
\title{Visually Imperceptible Adversarial Patch Attacks on Digital Images}
%
%
%

\author{Yaguan~Qian,
        Jiamin~Wang,
        Bin~Wang,
        Shaoning~Zeng,
        Zhaoquan~Gu,
        Shouling~Ji,
        and~Wassim Swaileh
\thanks{Y. Qian, J. Wang are with the School of Big-Data Science, Zhejiang University of Science and Technology, Hangzhou 310023, China. (e-mail: qianyaguan@zust.edu.cn)}
\thanks{B. Wang is with the Network and Information Security Laboratory of Hangzhou Hikvision Digital Technology Co., Ltd. Hang Zhou 310052, China. (e-mail: wbin2006@gmail.com)}
\thanks{S. Zeng is with the Yangtze Delta Region Institute, University of Electronic Science and Technology of China, Huzhou 313000, China. (e-mail:zsn@outlook.com)}
\thanks{Z. Gu is with the Cyberspace Institute of Advanced Technology, Guangzhou University, Guangzhou 510006, China. (e-mail: zqgu@gzhu.edu.cn)}
\thanks{S. Ji is with the College of Computer Science, Zhejiang University, Hangzhou 310023, China. (e-mail: sji@zju.edu.cn)}
\thanks{W. Swaileh is with the Computer Science Department of CY Cergy Paris University, Rouen, France. (e-mail: wassim.swaileh@litislab.fr)}
}
%
%

\markboth{Journal of \LaTeX\ Class Files,~Vol.~14, No.~8, August~2015}%
{Shell \MakeLowercase{\textit{et al.}}: Bare Demo of IEEEtran.cls for IEEE Communications Society Journals}
%



\maketitle

\begin{abstract}
The vulnerability of deep neural networks (DNNs) to adversarial examples has attracted more attention. Many algorithms have been proposed to craft powerful adversarial examples. However, most of these algorithms modified the global or local region of pixels without taking network explanations into account. Hence, the perturbations are redundant and they are easily detected by human eyes. In this paper, we propose a novel method to generate local region perturbations. The main idea is to find a contributing feature region (CFR) of an image by simulating the human attention mechanism and then add perturbations to CFR. Furthermore, a soft mask matrix is designed on the basis of activation map to  finely represent the contributions of each pixel in CFR. With this soft mask, we develop a new loss function with inverse temperature to search for optimal perturbations in CFR.  Due to the network explanations, the perturbations added to CFR are more effective than those added to other regions. Extensive experiments conducted on CIFAR-10 and ILSVRC2012 demonstrate the effectiveness of the proposed method, including attack success rate, imperceptibility, and transferability. 
\end{abstract}


%
\IEEEpeerreviewmaketitle

\section{Introduction}
%
%
%
%
\IEEEPARstart{T}{he} development of deep learning technology has promoted the successful application of deep neural networks (DNNs) in various fields, such as image classification \cite{[c10],[c11]}, computer vision \cite{[c12],[d1]}, natural language processing \cite{[c13],[d2]}, etc. In particular, convolutional neural networks (CNNs), as one of typical DNNs, have shown perfect performance in image classification. However, much evidence showed that CNNs are extremely vulnerable to \textit{adversarial examples} \cite{[a1]}. An adversarial example is crafted from a clean example by adding well-designed perturbations that are almost imperceptible to human eyes but can easily fool a classifier.  Though adversarial examples will lead misclassification, it as well provides a deep insight into the behaviors of CNNs \cite{[m]}. Goodfellow et al. \cite{[m]} argued that the primary cause of the adversarial instability is the linear nature and the high dimensionality of CNNs. Later work \cite{[c15]} studied the linearity hypothesis further and argued that adversarial examples exist when the classiﬁcation boundaries lie close to the manifold of sampled data. D. Su et al. \cite{su2018robustness} empirically found out the trade-off between accuracy and robustness and revealed that the robustness may be at the cost of accuracy.

\begin{figure}
\setlength{\abovecaptionskip}{-0.2cm}   
\setlength{\belowcaptionskip}{-0.5cm}    
\begin{center}
\includegraphics[width=9cm,height=3cm]{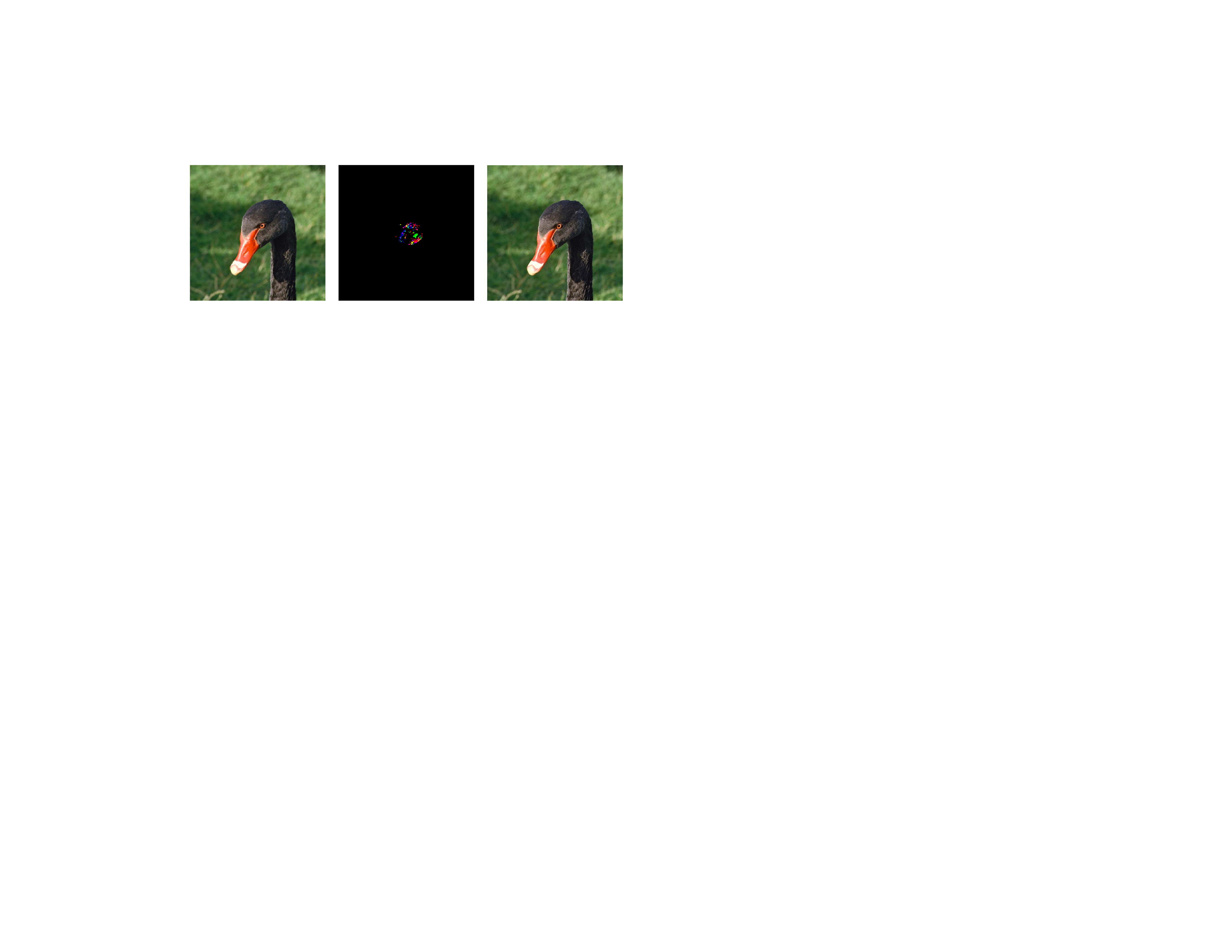} 
\end{center}
\caption{Left: the original natural image predicted as “black swan” with 75\% confidence. Middle: adversarial patch noise. Right: the adversarial image misclassified as “crampfish” with 70\% confidence, in which we can not observe the adversarial patch.}
\label{figa1}
\end{figure}

Recently, a variety of methods were proposed to craft adversarial examples, such as L-BFGS \cite{[a1]}, FGSM \cite{[m]}, I-FGSM \cite{[a3]}, PGD \cite{[a4]}, and C\&W \cite{[h]}, etc. These methods perturbed all the pixels in an image. At the same time, some methods only perturbed several pixels or a region of the image, such as JSMA \cite{[c3]}, One-pixel \cite{[c2]}, Adversarial Patch \cite{[c24]} and LaVAN \cite{[c25]}. However, since these techniques directly perturbed all pixels without considering semantic information, too many redundant perturbations are introduced in the irrelevant regions, e.g., back-ground. On the contrary, we attempt to generate more effective adversarial examples at semantic levels --- imperceptible adversarial patch, as shown in Fig. \ref{figa1}. 

Our method attempts to utilize the state-of-the-art explanation  \cite{[a12],[a11]} of CNNs to locate a proper perturbed region. Although the intrinsic mechanisms of CNNs are not fully understood by humans, some recent works have demonstrated more interesting clues \cite{[a12],[a11],[a13]}. These state-of-the-art explanations inspire us to craft local region perturbations in an interpretable way. Especially, the “attention mechanism” \cite{[c14]} inspires us to believe that not every region in an image makes the same contribution to the classification of CNNs, which is confirmed by \cite{[a12]} and  \cite{[a11]}. Therefore, if we find a sensitive region for classification and add perturbations to the region, it will be more effective to fool the classifier with fewer perturbations than previous methods \cite{[k]}. This special region is considered as a subspace of input feature space, which is referred to as a \textit{contributing feature region} (CFR) in this paper.  Specifically, we design a soft mask matrix to represent CFR for finely characterizing the contributions of each pixel. Based on this soft mask, we develop a new objective function to search for optimal perturbations in CFR. 

Although some local perturbations were proposed like JSMA \cite{[c3]} and One-pixel attacks \cite{[c2]}, they did not take account of the correlation between the neighboring pixels, i.e., the perturbed pixels may not form a continuous region. Meanwhile, some local continuous region attacks were proposed, which are named as patch attacks \cite{[c24],[c25]} or sticker attacks \cite{[c23]}. Our method also belongs to patch attacks. However, our method is different from those existing patch attacks in three aspects. The first is that the shape of patches or stickers is regular, while our CFR’s shape is arbitrary. The second is that the location of patches or stickers is optionally determined by the adversary, while our CFR is located by the explanation of CNNs at a semantic level. Finally, the perturbation magnitude of patches and stickers is not constrained, while our method limits the perturbations within a tiny bound to be imperceptible to human eyes. 

Furthermore, our work is different from image semantic segmentation \cite{[d3]} because our method pays more attention to the regions contributing to classification, while image segmentation is to find the object edge. In other words, we start with the network explanation while image segmentation focuses on the object itself. Recently, C. Xie et al. \cite{[c22]} crafted adversarial examples to fool semantic segmentation and object detection, while Z. Gu et al. \cite{[c16]} leveraged the YOLO detector to locate regions sensitive to the added perturbations. Essentially, their methods did not fully utilize the network explanation. Besides, the size of our CFR is far smaller than the region obtained by object detectors.  W. Wu et al. \cite{wu2020boosting} utilize the attention mechanism to investigate the transferability of adversarial examples across multi-models in black-box settings, while our goal is to improve the attack success rate with imperceptible patch perturbations in both white-box and black-box settings. 

The main contributions can be summarized as follows:
\begin{itemize}
\item We propose a visually imperceptible adversarial patch attack, which combines network explanations and optimization techniques to achieve a good tradeoff between intensity and imperceptibility. Compared to previous patch attacks, the advantage of our method lies in two aspects. First, the magnitude of perturbations is substantially reduced, which is almost imperceptible to humans. Second, the patch position is optimized by the attention mechanism at a semantic level.
\item Our work shows that the adversarial patch located by network explanations can effectively fool CNNs, which reveals that the CNN has an attention mechanism similar to humans. It provides a crucial clue for exploration of effective countermeasure against adversarial examples in the future.
\item Extensive experiments are conducted on CIFAR-10 and ILSVRC2012, which demonstrate that CFR patch attack consistently outperforms state-of-the-art methods no matter in white-box or black-box settings. In brief, compared with the recently proposed patch attacks, the patch crafted by our method is imperceptible; among all the global attacks, the magnitude of our perturbations is the smallest. 
\end{itemize}

The rest of the paper is organized as follows. The next section highlights the related works in various adversarial attacks, including global perturbations and local perturbations at pixel levels. Then we describe our method to craft perturbations on CFRs in Section \ref{sec3}, and the experimental results are shown in Section \ref{sec4}. Finally, we conclude the paper in Section \ref{sec5}.



%


\section{Related Work}
In this section, we introduce some related work in adversarial attacks according to the perturbed regions. These attacks are divided into global adversarial attacks and local adversarial attacks. Our approach proposed in this paper belongs to the latter. However, different from previous local attacks at the pixel level, we add perturbations to some semantic regions through network explanations. The global adversarial attacks at the pixel level are introduced in Section \ref{sec2.1}, and the local adversarial attacks at the pixel level are presented in Section \ref{sec2.2}.

\subsection{Global Adversarial Attacks at Pixel Levels}
\label{sec2.1}
C. Szegedy et al. \cite{[a1]} first found that DNNs are vulnerable to adversarial examples, and they proposed a box-constrained optimal perturbation method called L-BFGS. Since L-BFGS used an expensive linear search method to find the optimal perturbation, it was time-consuming and impractical. I. J. Goodfellow et al. \cite{[m]} proposed FGSM (Fast Gradient Sign Method) to generate adversarial examples. This method only performed an one-step gradient update along the direction of the sign of gradient at each pixel, so the computation cost was extremely lower. However, the generated adversarial example may not have the best performance since it is only a roughly approximation. As shown in \cite{[a3]}, an one-step attack like FGSM is easy to transfer across multi-models but also defensilbe. On the basis of FGSM, many other improved methods were proposed, such as I-FGSM \cite{[c17]}, in which iteration was applied to FGSM to generate adversarial examples. F. Tramèr et al. \cite{[c18]} found that FGSM with adversarial training was more robust to white-box attacks than black-box attacks due to gradient masking. They proposed a RAND-FGSM, which added random noise when updating the adversarial examples to defeat adversarial training. Y. Dong et al. \cite{[c19]} proposed MI-FGSM, assuming that the gradients of each iteration were not only related to the current gradients, but also related to the gradients of the previous iteration. More recently, M-DI$^{2}$-FGSM \cite{[c8]} based on MI-FGSM was proposed to improve transferability in black-box attacks.

Besides these FGSM series of attacks, a variety of other improved algorithms have been proposed. S.-M. Moosavi-Dezfooli et al. \cite{[c20]} proposed Deepfool to find the closest distance from the original input to the decision boundary of adversarial examples. To overcome the non-linearity in the high dimension space, they performed an iterative attack with a linear approximation. DeepFool provided less perturbation compared with FGSM. N. Carlini and D. Wagner \cite{[h]} proposed the C\&W method to defeat defensive distillation. C\&W considered three forms of perturbation constraints ($\ell_{0}$, $\ell_{1}$ and $\ell_{\infty}$ norm of the added perturbations) and adjusted the added perturbations with the optimization method. As far as we know, C\&W is one of the most powerful attacks at the pixel level. All of these works achieve some significant progress, however, they do not fully take into account the semantic of images, which will bring more redundant perturbations to pixels.
\vspace{-0.369cm}

\subsection{Local Adversarial Attacks at Pixel Levels}
\label{sec2.2}
Different from manipulating each image pixel for misclassification, several methods are proposed to perturb multiple pixels, which are called local adversarial attacks in this paper. A classic method is JSMA proposed by N. Papernot et al. \cite{[c3]} for targeted attacks. They perturbed a small number of pixels by a constant offset in each iteration step that maximizes the saliency map. However, JSMA has the disadvantage of over-modifying the value of the pixels, making the added perturbations easily perceived by the naked eye, and its adversarial strength is weak \cite{[c1]}. J. Su et al. \cite{[c2]} proposed the One-pixel attack method that successfully deceived the DNNs by modifying the value of a single pixel. Although this method is better for low-resolution images (such as CIFAR-10), the attack success rate for high-resolution images will be greatly reduced (such as ImageNet), and the cost is very large with the $\ell_{1}$ distortion \cite{[c4]}. 

While the pixels perturbed by JSMA are usually nonadjacent, while in another method the pixels perturbed are in a continuous region. I. Evtimov et al. \cite{[c23]} proposed a sticker attack that added noise patches as rectangular patterns on the top of trafﬁc signs. T. B. Brown et al. \cite{[c24]} presented a method called Adversarial Patch to create universal, robust, targeted adversarial image patches in the real world. Our method is different from them in three aspects. First, the size of the patch in Adversarial Patch is determined manually while in our method it is determined by CFR automatically. Second, the patch location of Adversarial Patch can be random in the image, while the position in our method is located through Grad-CAM \cite{[a12]} at the semantic level. Finally, the patches are recognizable in Adversarial Patch, while our patches are imperceptible. D. Karmon et al. \cite{[c25]} suggested LaVAN that generate localized adversarial noises that cover only 2\% of the pixels in the image, none of them cover the main object. Besides, it is transferable across images and locations, and can successfully fool a state-of-the-art Inception v3 model with very high success rates. Although these stickers and patch attacks have demonstrated powerful strength and can easily bypass existing defense approaches, there is no constraint on noise and they can be observed easily. Moreover, the location of stickers or patches is randomly determined by the adversary. Z. Gu et al. \cite{[c16]} leveraged the YOLO detector to locate sensitive regions for perturbations. In contrast, the goal of our method is to locate a sensitive region through network explanations and add imperceptible perturbations. 

\section{Methodology}
\label{sec3}
\subsection{Preliminary}
A CNN can be generally expressed as a mapping function $f(X, \theta): \mathbb{R}^{m} \rightarrow \mathbb{R}^{C}$, where $X \in \mathbb{R}^{m}$ is an input variable, $\theta$ denotes all the parameters and $C$  is the number of classes. Typically, A CNN is comprised of convolutional layers with some method of periodic downsampling (either through pooling or stride convolutions). Let $Z$ be the output vector of the penultimate layer,  namely the Logits layer. This defines a mapping function: $X \mapsto Z$.  The last layer of the CNN is a softmax layer. Then the softmax function can be expressed as $S(Z)_{j}=\exp(Z_{j}) / \sum_{i=1}^{C} \exp (Z_{i})$, where $Z_i$ is the $i$-$th$ element of $Z$, and $i \in [C]$, $[C]=\{1,...,C\}$ is a set of class labels. Thus the CNN can be expressed as $f(X)=S\left(W_{s} Z+b_{s}\right)$, where $W_{s}$ and $b_{s}$ are the weight matrix and bias vector of the softmax layer respectively. Given an input $X$ with the ground-truth class label $y$, the predicted class label of $X$ can be expressed as $\hat{y}=\arg \max _{i\in [C]} f(X)_{i}$. An adversarial example can be represented as $X^{\prime}=X+\delta$, where $X$ is a clean nature image, and $\delta$ is the perturbation. To obtain imperceptible perturbations, $\delta$ is always constrained by a $p$-norm, $\|\delta\|_{p} \leq \epsilon$, where $p=\ell_{0}, \ell_{2}$ or $\ell_{\infty}$ and $\epsilon$ is the perturbation bound.

\subsection{Threat Model}
In general, the method used to generate adversarial examples needs some proper assumptions. These assumptions consist of a so-called threat model. X. Yuan et al. \cite{[c26]} presented a deep learning threat model in two dimensions. The first dimension is the adversarial goal including \textit{targeted attacks} and \textit{untargeted attacks} according to the adversarial specificity. The second dimension is the attacker ability defined by the amount of information that the attacker can obtain from target CNNs, which are divided into two categories, i.e., \textit{white-box} attacks and \textit{black-box} attacks.

\textbf{Adversarial Goal:} For untargeted attacks, the adversarial example $X^{\prime}$ satisfies $y^{\prime} \neq y$, i.e., $y^{\prime}$ can be any class label except $y$ (the ground-truth label of $X$), where $y^{\prime}=\arg \max _{i \in [C]} f\left(X^{\prime}\right)_{i}$. For targeted attacks, we specify a target class label $y^{*}$ , and the adversarial example $X^{\prime}$ must satisfy $y^{*}=\arg \max _{i \in [C]} f\left(X^{\prime}\right)_{i}$ and $y^{*} \neq y$. In this paper, we mainly focus on untargeted attacks that are suitable for further adversarial training as a countermeasure.

\textbf{Adversarial Capabilities} are defined by the amount of information that the adversary has about the target classifier. The so-called white-box attack means that the adversary has almost all the information about the target CNN, including training data, activation functions, network topologies, and so on. The black-box attack, however, assumes that the attacker has no way to access the internal information of the pretrained CNN, except the output of the model including the label and confidence. In this paper, we assume white-box settings as the same as PGD, C\&W, etc. Nevertheless, all white-box attacks can be employed to lunch black-box attacks through a substitute model as described in \cite{[c27]}.

\subsection{Problem Formulation}
In this paper, we attempt to add perturbations to a local region instead of the whole image. To formalize the problem of this patch attack, let $X_{r}$ denote the region to which the perturbation $\delta_{r}$ is added. Note that $X_{r}$  is generally not a regular region. For the convenience of calculation, we introduce a binary matrix $M$ to represent the shape of $X_{r}$. Here $M$ is, in general, a 0-1 matrix:

\begin{equation}
\label{eq1}
M(i, j)=\left\{\begin{array}{ll}
1 & X(i, j) \in X_{r}, \\
0 & \text { otherwise },
\end{array}\right.
\end{equation}
where $X(i, j)$ is a pixel at the cell $(i,j)$ of $X$. Thus we can transfer $X_{r}$ to a matrix by $X \odot M$, where $\odot$ is the Hadmard product. Accordingly, $\delta_{r}$ can be represented by $\delta \odot M$, where $\delta$ is a global perturbation. Thus, obtaining an optimal $\delta_{r}$ can be modeled as the following constrained optimization problem:
 \begin{alignat}{2}
 \label{eq2}
     \min  \quad & \|\delta \odot M\|_{p}, & \\
     \mathrm{s.t.} \quad & f(X+\delta \odot M) \neq y, \, 
         \nonumber \\
   & X+\delta \odot M \in[0,1]^{m} \,  \nonumber .
 \end{alignat}
 
However, solving the problem (\ref{eq2})  is non-trivial. Instead, we obtain perturbations by maximizing the loss function as doing in most previous work: 
  \begin{alignat}{2}
 \label{eq3}
     \max  \quad & J\left(X+\delta \odot M, y\right),& \\
     \mathrm{s.t.} \quad & \ X+\delta \odot M \in[0,1]^{m} \,  \nonumber ,
 \end{alignat}
where $J$ is a loss function. Although problem  (\ref{eq3}) is not fully equivalent to (\ref{eq2}) and thus may not guarantee all obtained perturbations to flip the class label, its advantage is to fast find a possible perturbation within the constrained range like FGSM. We think solving problem (\ref{eq3}) depends on two aspects. The one is to locate $X_{r}$, i.e., identifying the binary matrix $M$, as described in Section \ref{sec3.3}; the other is to find a proper loss function to solve problem (\ref{eq3}), as demonstrated in Section \ref{sec3.4}. 

\subsection{Contributing Feature Regions (CFRs)}
\label{sec3.3}
Suppose the input image $X\in \mathbb{R}^{m}$  is forward propagated through the CNN, and the final convolutional layer outputs the high-level feature map $A$ of the image, where $A^{(k)} \in \mathbb{R}^{u \times v}$ represents the feature map of the $k$-$th$ convolutional kernel with the size of $u \times v$. Next, $A$ passes through the fully connected layers and finally outputs a confidence vector $Z$. Let $Z_{c}$ represent the logits of the $c$-$th$ class. A larger value of $Z_{c}$ indicates $X$ is predicted to the $c$-$th$ class with a greater probability. To this end, we compute the gradient of $Z_{c}$ with respect to $A^{(k)}$, i.e., $\partial Z_{c} / \partial A^{(k)}$ to measure the classification prediction importance of the $k$-$th$ convolutional kernel to the $c$-$th$ class. Furthermore, we adopt the global average pooling operation to calculate the weight $\lambda_{c}^{(k)}$ of the $k$-$th$ convolutional kernel:

\begin{equation}
\lambda_{c}^{(k)}=\frac{1}{u\times v} \sum_{p} \sum_{q} \frac{\partial Z_{c}}{\partial A_{p q}^{(k)}},
\end{equation}
where $A_{p q}^{(k)}$ is the activation at the cell $(p, q)$ of the $k$-$th$ convolutional kernel. Thus, we obtain a feature activation map $\sum_{k} \lambda_{c}^{(k)} A^{(k)}$ for the $c$-$th$ class. Considering that only the positive elements in  $\sum_{k} \lambda_{c}^{(k)} A^{(k)}$ have a positive effect on the classification, the result is further reactivated by ReLU to remove the influence of negative elements, and the final activation map of the  $c$-$th$ class is obtained:

\begin{equation}
L_{c}=\operatorname{ReLU } \left(\sum_{k} \lambda_{c}^{(k)} A^{(k)}\right).
\end{equation}

In fact, in our work the $c$-$th$ class is the ground-truth class label $y$ of $X$. Then, we substitute $y$ for $c$. For further distinguishing the contribution of each pixel in CFR, we design a soft-mask $\tilde {M}$ instead of using the traditional binary hard-mask in Eq. (\ref{eq1}):
\begin{equation}
\tilde {M}(i, j)=\left\{\begin{array}{l}
\frac{L_{y}(i, j)}{ \sum\limits_{m}\sum\limits_{n} L_{y}(m, n)} \quad L_{y}(i, j) \geq \tau, L_{y}(m, n) \geq \tau, \\
0 \quad \quad \quad \quad \quad     \text { otherwise },
\end{array}\right.
\end{equation}
where $\tau$ is a threshold. 

\subsection{Generate Perturbations for CFRs}
\label{sec3.4}
After locating CFR, we further generate the local perturbation $\delta_{CFR}$ on CFR. We design a new loss function to implement problem (\ref {eq3}), which consists of two parts: (1) a cross-entropy loss function $J_{CE}$ for generating adversarial examples, and (2) an $\ell_{2}$ regularization function to restrict the perturbation:
\begin{equation}
\label{eq7}
J=J_{CE}+\beta \frac{1}{\left\|\delta \odot \tilde {M}\right\|_{2}},
\end{equation}
where $\beta$ is a hyper-parameter to control the degree of distortion (we set $\beta=1$) and $\delta \odot \tilde {M}$ represents $\delta_{CFR}$. In theory, $\ell_{0}$ or $\ell_{\infty}$ can also be used for regularization. However, we notice that the $\ell_{0}$ norm is non-differentiable and hard to caculate for the standard gradient descent algorithm. Besides, the $\ell_{\infty}$ norm only focuses on the largest value in $\delta_{CFR}$, it easily fluctuates between two sub-optimal solutions during the gradient descent process \cite{[h]}.

Remind the original cross-entropy loss function $J_{CE}=-\log S_{y}$, where $S_{y}=\exp(Z_{y}) / \sum_{i=1}^{C} \exp(Z_{i})$.  In adversarial settings, we aim to maximize $J_{CE}$ to obtain an adversarial example. However, when $S_{y}$ tends to approach $1$, $J_{CE}$ is close to $0$. Thus, the update of $\delta_{CFR}$ has minimal impact on $J_{CE}$, which is undesirable to us. To avoid this situation, we introduce a hyper-parameter $T$ ($T>0$) called \textit{inverse temperature} inspired by the distillation idea \cite{[c9]}. In \cite{[c9]}, they leverage $T$ to smooth the confidence distribution of classes, while we use it to maintain the impact of loss during the back-propagation. Then $J_{CE}$ is modified as follows:

\begin{equation}
J_{CE}^{\prime}=\frac{-\log \left(S_{y}\right)}{T},
\end{equation}
where $S_{y} \in(0,1)$ and $\log \left(S_{y}\right) \in(-\infty, 0)$. If $0<T<1$, the lower bound of $\log \left(S_{y}\right) / T$ is magnified and $-\log \left(S_{y}\right) / T$ becomes larger, that is, $J_{CE}^{\prime}$ becomes larger. If $T>1$, the lower bound of $\log \left(S_{y}\right) / T$ is reduced and $-\log \left(S_{y}\right) / T$ becomes smaller, that is, $J_{CE}^{\prime}$ gets smaller. Our goal is to maximize $J_{CE}^{\prime}$, so we set $0<T<1$. Thus we redefine problem (\ref{eq7}) as follows:

 \begin{alignat}{2}
 \label{eq8}
     \max  \quad &J_{CE}^{\prime}+\beta \frac{1}{\left\|\delta \odot \tilde {M}\right\|_{2}}, & \\
     \mathrm{s.t.} \quad &X+\delta \odot \tilde{M }\in[0,1]^{m}  \, 
         \nonumber .
  \end{alignat}
Finally, we use the hill climbing algorithm to solve problem (\ref{eq8}) as shown in Algorithm 1.

\begin{algorithm}
	\caption{Crafting Adversarial Patch Examples}
     
    {\textbf{Input:} A clean image $(X,y)$, the iterations $N$, step size $\eta$, degree of distortion $\beta$, threshold $\tau$, and inverse temperature $T$}\\
    {\textbf{Output:} An adversarial example $X^{\prime}$}

\begin{algorithmic}[1]

		\STATE initialize $\delta$ \\ 
		// $K$ is the number of feature maps in the last layer of convolution layers \\
		\STATE $\lambda_{y}^{(k)} \leftarrow \frac{1}{u \times v} \sum_{p} \sum_{q} {\partial Z_{y}}/{\partial A_{p q}^{(k)}}$, $k=1 \ldots K$ \\
		\STATE  $L_{y} \leftarrow \operatorname{ReLU}\left(\sum_{k} \lambda_{y}^{(k)} A^{(k)}\right)$ \\
		 // Get a CFR, $i=0 \ldots u$, $j=0 \ldots v$  
		\STATE \textbf{if} $L_{y}(i, j) \geq \tau$ \textbf{and} $L_{y}(m, n) \geq \tau$ \textbf{then} \\
		\STATE \qquad $\tilde{M}(i, j) \leftarrow {L_{y}(i, j)}/{\sum_{m} \sum_{n} L_{y}(m, n)}$ \\
		\STATE  \textbf{else} \\
		 \STATE  \qquad  $\tilde{M}(i, j) \leftarrow 0$ \\
		 \STATE \textbf{end if}
		\STATE \textbf{for}  $t=1 \ldots N$ \textbf{do}\\
		\STATE  \qquad $ J \leftarrow J_{CE}^{\prime}+\beta / {\left\|\delta_{t} \odot \tilde {M}\right\|_{2}}$\\
		 // Update $\delta$
		\STATE  \qquad $\delta_{t+1} \leftarrow\left(\delta_{t}+\nabla_{\delta_{t}}J\times \eta \right) \odot  \tilde {M}$   \\
		\STATE \qquad $X^{\prime}_{t+1} \leftarrow \text{Clip}(X+\delta_{t+1}, 0, 1)$ \\
		\STATE  \textbf{end for}\\
\end{algorithmic}
\end{algorithm}

\section{Experiments}
\label{sec4}
In this section, we first describe the datasets, models, and metrics used in our experiment. Then we show the impact of CFR on classification by simply setting CFR to 0. The results confirm that CFR plays a critical role in classification. For intuition, we visualize the adversarial examples and their perturbations of various methods. In white-box settings, we use ASR, SSIM, and $\ell_{p}$ to make a comparison between our method and the classic global attacks such as PGD and C\&W, and local attacks such as JSMA, One-pixel. In addition, two recent patch attacks Adversarial Patch and LaVAN are aslo compared with our method. Considering the above experiments are conducted on non-protected models, we further investigate the effect of our method on protected models. Furthermore, we compare the transferability of our method with other attacks in black-box settings. Finally, several key hyper-parameters are discussed in this paper.

\subsection{Experiment Setup}
\textbf{Datasets and Models.} We validate our method on two benchmark datasets CIFAR-10 \cite{[a20]} and ILSVRC2012 \cite{[a17]}. CIFAR-10 consists of 60,000 images with the size of $32 \times 32$, including 10 categories and each with 6,000 images, in which 50,000 images are used for training and 10,000 images for tests. ILSVRC2012 contains 1,000 categories, in which 1,200 thousand images are used for training, and 50,000 images for tests. All the images we use to generate adversarial examples are correctly classified by all models, which can guarantee all the misclassified examples are adversarial examples. Two popular CNNs VGG \cite{[c11]} and ResNet \cite{[c12]} are selected for our experiment. According to their number of layers, they are further divided into VGG-11, VGG-13, VGG-16, ResNet-18, ResNet-34, and ResNet-50. 

\textbf{Evaluation Metrics.} We use (1) the attack success rate (ASR) to measure the power of the adversarial examples, (2) the $\ell_{p}$ norm to measure the perturbation amplitude, and (3) the structural similarity (SSIM) index as a measurement of image similarity because human visual perception is highly sensitive to the structural information of an image \cite{[d9]}.

(1) \textbf{ASR:} Given $n$ clean images correctly classified by a CNN, the corresponding adversarial examples are obtained by a special generating method. Suppose $X_{i}$ represents the $i$-$th$ clean image, its ground-truth label is $Y_{i}$, and $X_{i}^{\prime}$ is its corresponding adversarial image. Then ASR can be obtained by the following formula:

\begin{equation}
ASR=100 \times \frac{1}{n} \sum_{i=1}^{n} \mathbb{I}\left[\arg \max _{j\in [C]} f\left(X_{i}^{\prime}\right)_{j} \neq Y_{i}\right],
\end{equation}
\vspace{0.15cm}
where $\mathbb{I}[\cdot]$ is an indicator function and $[C]=\{1,...,C\}$ is a set of class labels.

(2) \textbf{SSIM:} Given a clean image $X$ and its corresponding adversarial image $X^{\prime}$, $\operatorname{SSTM}\left(X, X^{\prime}\right)$ measures the similarity between $X$ and $X^{\prime}$. A larger $\operatorname{SSIM}\left(X, X^{\prime}\right)$ indicates a higher similarity between the two images.

\begin{equation}
\operatorname{SSIM}\left(X, X^{\prime}\right)=\left[l\left(X, X^{\prime}\right)\right]^{\alpha}\left[c\left(X, X^{\prime}\right)\right]^{\beta}\left[s\left(X, X^{\prime}\right)\right]^{\gamma},
\end{equation}
where $\alpha$, $\beta$, $\gamma>0$, $l\left(X, X^{\prime}\right)$ is brightness comparison, $c\left(X, X^{\prime}\right)$ is contrast comparison, and $s\left(X, X^{\prime}\right)$ is structure comparison:

\begin{equation}
l\left(X, X^{\prime}\right)=\frac{2 \mu_{X} \mu_{X^{\prime}}+c_{1}}{\mu_{X}^{2}+\mu_{X^{\prime}}^{2}+c_{1}},
\end{equation}

\begin{equation}
c\left(X, X^{\prime}\right)=\frac{2 \sigma_{X X^{\prime}}+c_{2}}{\sigma_{X}^{2}+\sigma_{X^{\prime}}^{2}+c_{2}},
\end{equation}

\begin{equation}
s\left(X, X^{\prime}\right)=\frac{\sigma_{X X^{\prime}}+c_{3}}{\sigma_{X} \sigma_{X^{\prime}}+c_{3}},
\end{equation}
where $\mu_{X}$ and $\mu_{X^{\prime}}$ represent the average of $X$ and $X^{\prime}$ respectively, $\sigma_{X}$ and $\sigma_{X^{\prime}}$ represent the standard deviation of $X$ and $X^{\prime}$ respectively, $\sigma_{X X^{\prime}}$ represents the covariance of $X$ and $X^{\prime}$, and $c_{1}$, $c_{2}$, and $c_{3}$ are constants.

\subsection{Impact of CFR on Classification}
\label{IV-B}
\begin{figure}
\setlength{\abovecaptionskip}{-0.2cm}   
\setlength{\belowcaptionskip}{-0.8cm}    
\begin{center}
\includegraphics[width=7.8cm,height=4.8cm]{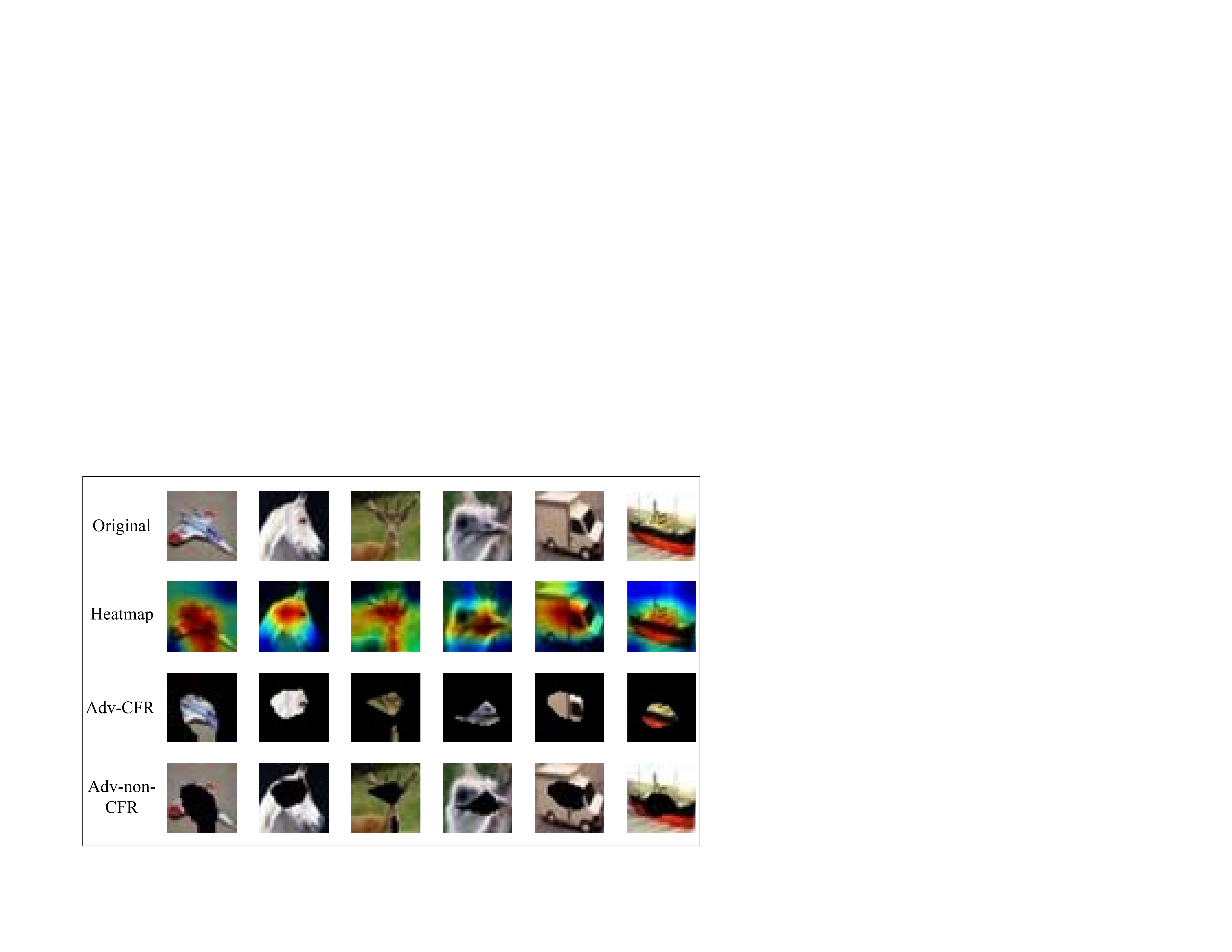} 
\end{center}
\caption{Two groups of special adversarial images: Adv-CFR and Adv-non-CFR.}
\label{figa2}
\end{figure}

\begin{figure}
\setlength{\abovecaptionskip}{-0.2cm}   
\setlength{\belowcaptionskip}{-0.5cm}    
\begin{center}
\includegraphics[width=7.5cm,height=4cm]{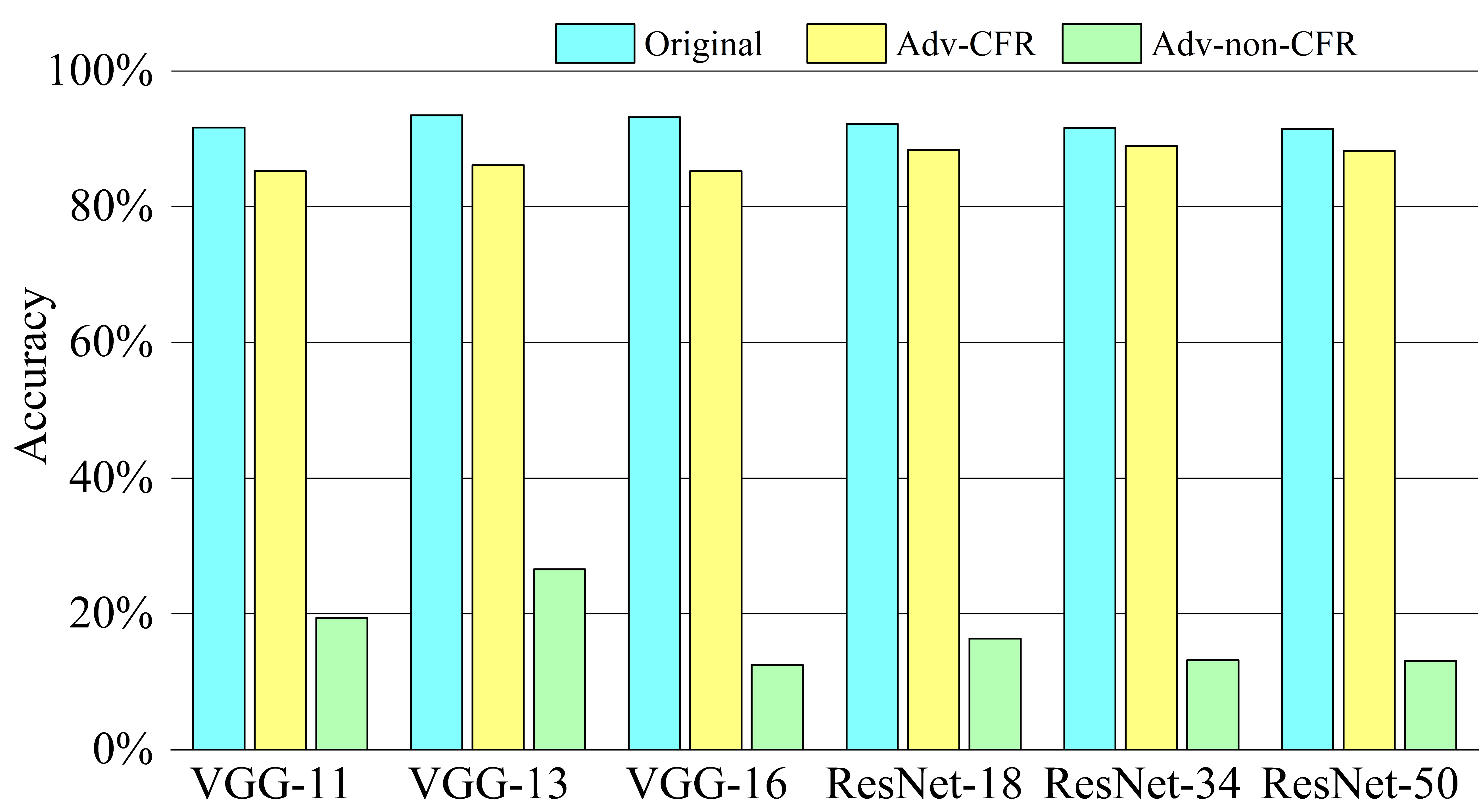} 
\end{center}
\caption{Accuracy of different models with clean images, Adv-CFR, and Adv-non-CFR examples}
\label{figa3}
\end{figure}

We first evaluate the impact of CFR on the classifier through two groups of special adversarial images. The images in one group keep the pixels in CFR unchanged while the rest of the pixels are set to 0, which is denoted as Adv-CFR. On the contrary, the images in the other group keep the pixels unchanged other than CFR that is set to 0, which is denoted as Adv-non-CFR. These special adversarial images are crafted from 10,000 clean images on CIFAR-10 and the threshold $ \tau=0.2$ is adopted. Fig. \ref{figa2} shows the examples of Adv-CFR and Adv-non-CFR. These examples of Adv-CFR and Adv-non-CFR are tested on VGG and ResNet, and the results are shown in Fig. \ref{figa3}. Compared to the original clean images, the accuracy of Adv-CFR decreases by no more than 3\%, however, the accuracy of Adv-non-CFR decreases by at least 60\%. The result shows that though the size of CFR is smaller than non-CFR, it yet plays a key role in classification. Therefore, the adversary modifying CFR is effective than modifying other regions to fool a classifier.

\begin{figure}
\setlength{\abovecaptionskip}{-0.2cm}   
\setlength{\belowcaptionskip}{-0.5cm}    
\begin{center}
\includegraphics[width=8cm,height=4.8cm]{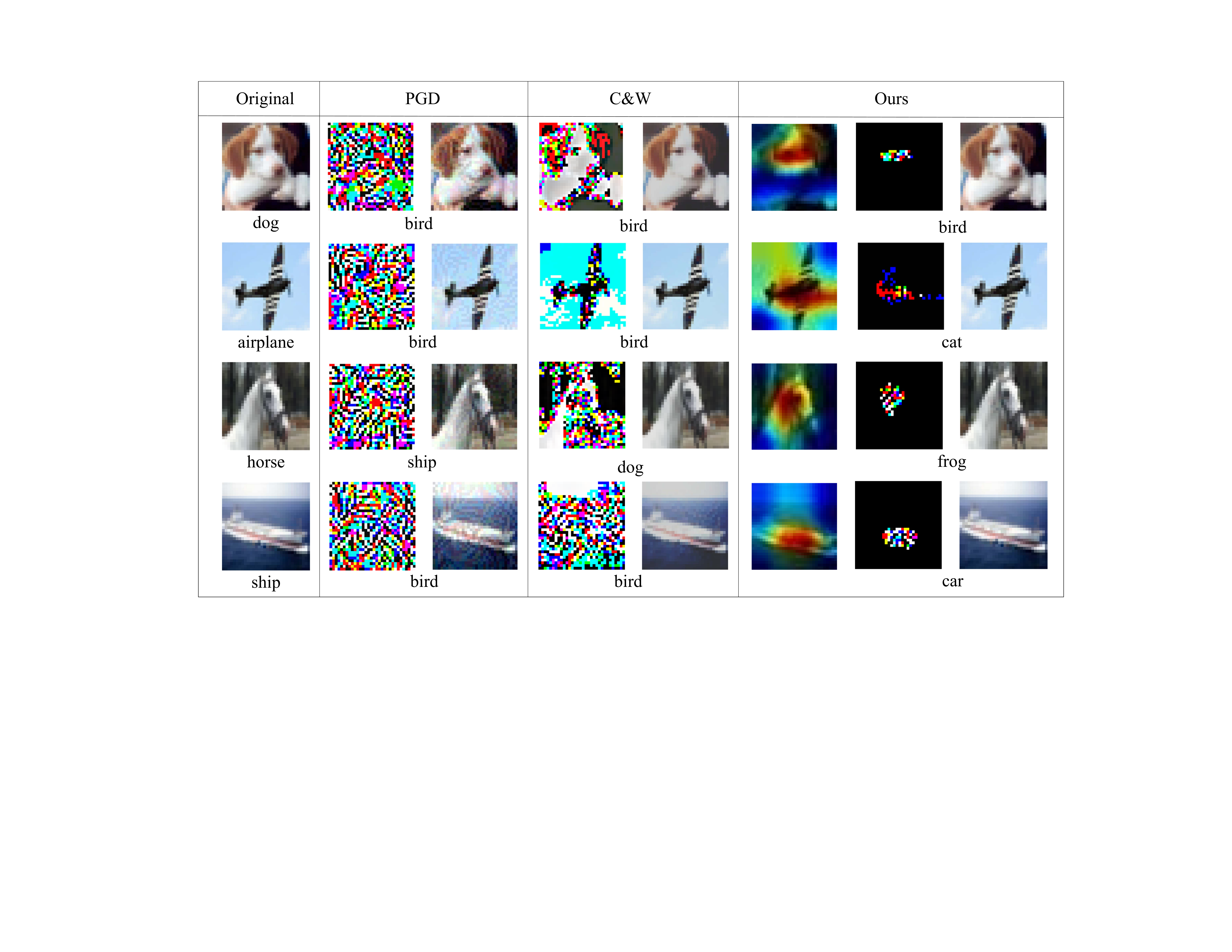} 
\end{center}
\caption{Comparison of perturbations and adversarial examples generated by PGD, C\&W, and our method on CIFAR-10 (all these three methods are constrained by $\ell_{2}=2$).}
\label{figa4}
\end{figure}

\subsection{Visualizing Adversarial Attacks}
\begin{figure*}
\setlength{\abovecaptionskip}{-0.2cm}   
\setlength{\belowcaptionskip}{-0.5cm}    
\begin{center}
\includegraphics[width=16cm,height=11cm]{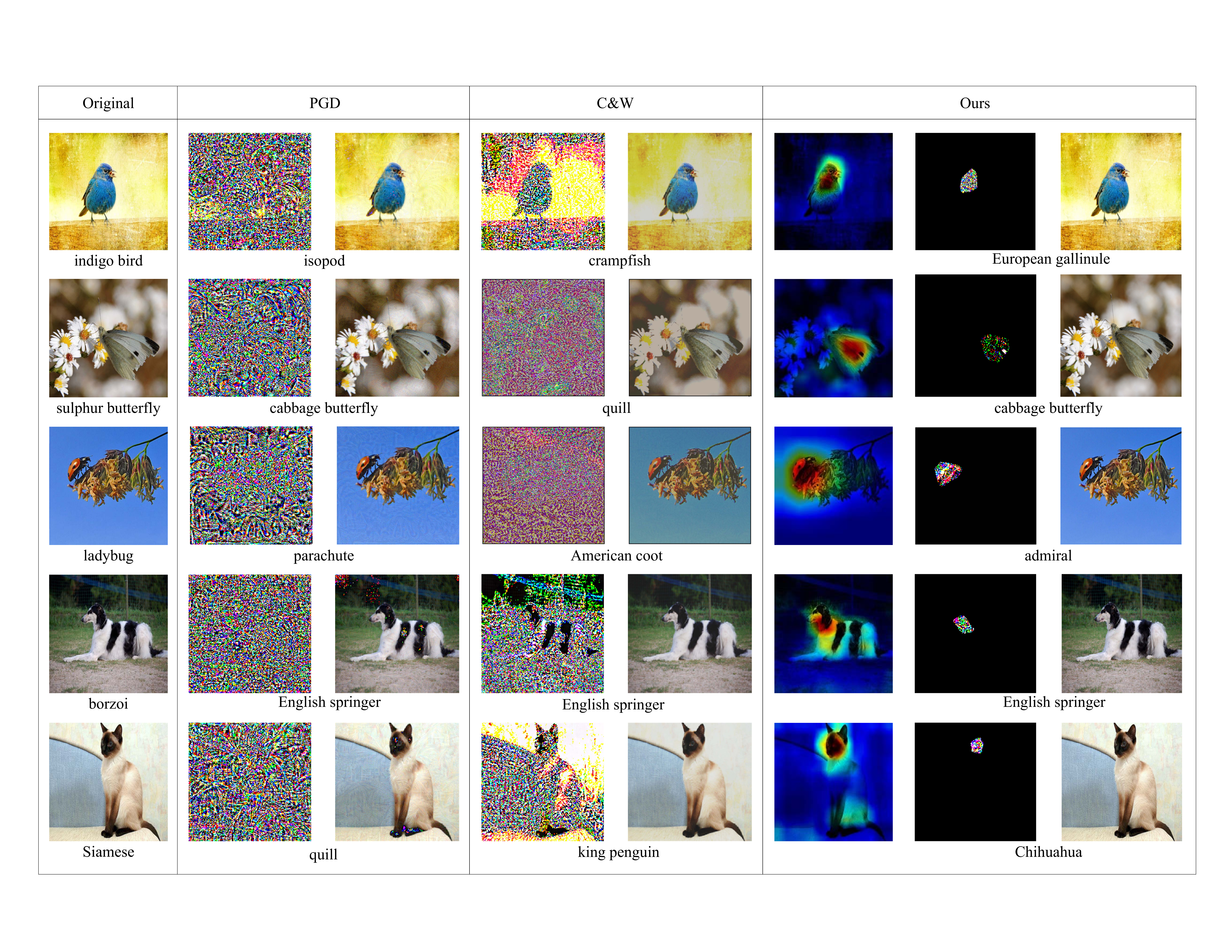} 
\end{center}
\caption{ Comparison of perturbations and adversarial examples generated by PGD, C\&W, and our method on ILSVRC2012 (all these three methods are constrained by $\ell_{2}=15$).}
\label{figa5}
\end{figure*}

\begin{figure*}
\setlength{\abovecaptionskip}{-0.2cm}   
\setlength{\belowcaptionskip}{-0.5cm}    
\begin{center}
\includegraphics[width=10cm,height=10.5cm]{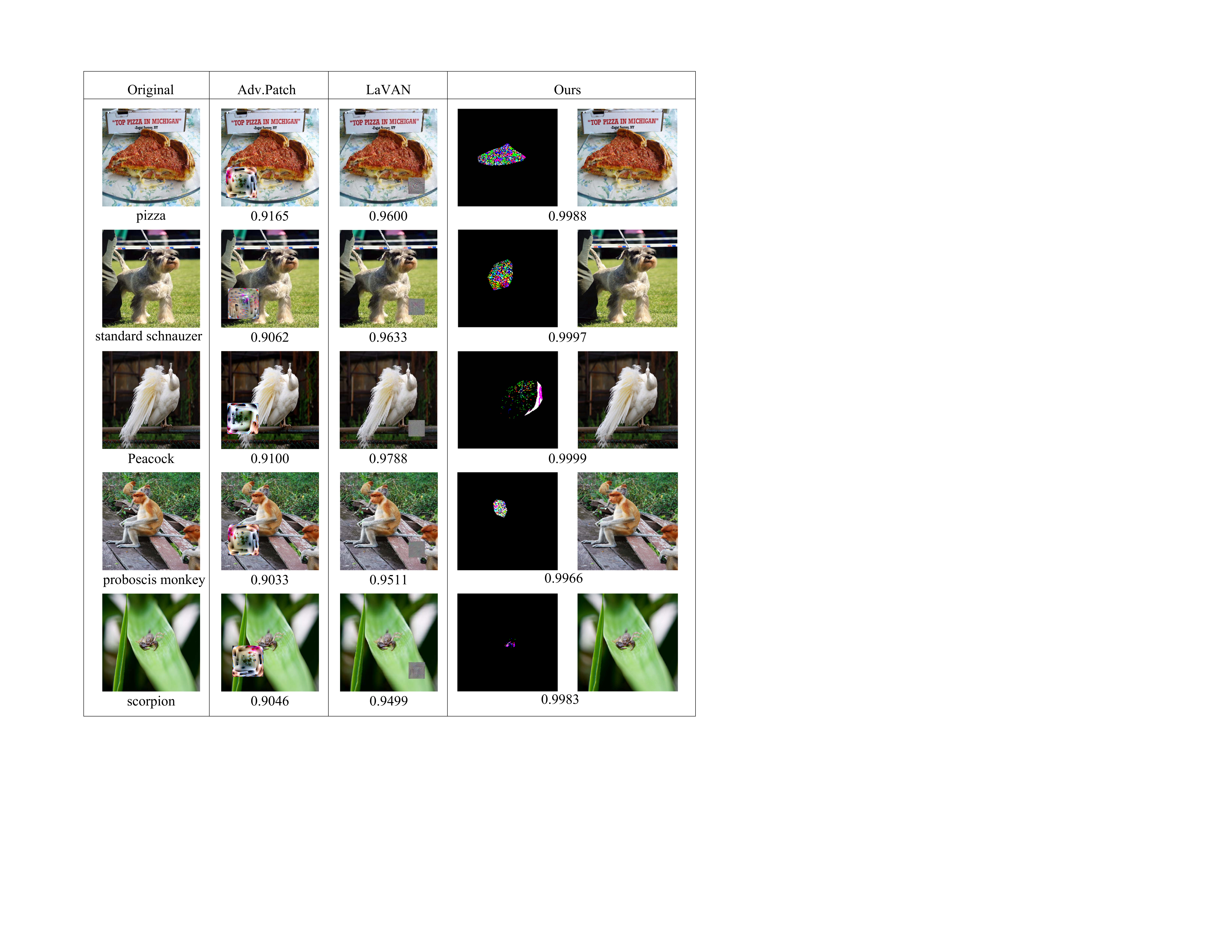} 
\end{center}
\caption{ SSIMs of three patch attacks on ILSVRC2012. The penultimate row visualizes the patch perturbation generated by our method.}
\label{figa6}
\end{figure*}

To intuitively illustrate the adversarial examples and their perturbations, we present them in Fig. \ref{figa4}, \ref{figa5}, and \ref{figa6}. We first compare our CFR patch with global pixel perturbations PGD and C\&W on CIFAR-10 (see Fig. \ref{figa4}) and ILSVRC2012 (see Fig. \ref{figa5}).  On CIFAR-10, the average SSIM of PGD is 0.94 and C\&W is 0.96, while our method is 0.99. On ILSVRC2012, the average SSIM of PGD is 0.91 and C\&W is 0.95, while our method is 0.99. Remind that a higher SSIM indicates a higher similarity between the images. It can be seen that the images with our patch perturbations are perfectly closer to the original clean images than those global perturbed images. 

Two recent patch attack methods—Adversarial Patch, denoted as Adv. Patch \cite{[c24]} and LaVAN \cite{[c25]} are compared with our method on ILSVRC2012. We do not conduct patch attacks on CIFAR-10 due to its low resolution and small size. For Adversarial Patch, we use the same experimental setting as \cite{[c24]} that the adversarial patch covers 10\% of pixels. For LaVAN, which is an improved version of Adversarial Patch, we follow the same implementation as \cite{[c25]} where the adversarial patch only covers 3\% pixels. Fig. \ref{figa6} presents the patches, adversarial examples, and SSIM of the above methods. Obviously, the patch crafted by Adv.Patch and LaVAN can be easily detected by humans. It can also be confirmed by SSIM that our method is ultimately close to 1, which is higher than Adv.Patch and LaVAN, and indicates the adversarial image crafted by our method is perfectly imperceptible to humans.

\begin{table*}
\caption{asr, ssim, and $\ell_{p}$ distortion of various attacks}
\setlength{\tabcolsep}{4mm}   
\label{table1}
\centering
\begin{tabular}{cccccccc}
\toprule  

Dataset                     & Attack Methods    & ASR      & SSIM & $\ell_{0}$  & $\ell_{1}$  & $\ell_{2}$  &$\ell_{\infty}$   \\ \hline
\multirow{5}{*}{CIFAR-10}   & PGD               & 93.18\%  & 0.94 & 3,060   & 144.79    & 2.85   & 0.06                 \\
                            & C\&W              & 97.44\%  & 0.96 & 3,072   & 18.11    & 0.49  & 0.05                 \\
                            & JSMA              & 90.33\%  & 0.71 & 335    & 856.11    & 28.04  & 1.00                 \\
                            & One-pixel         & 80.77\%  & 0.99 & 15     & 24.89     & 7.03   & 1.00                 \\
                            & Ours              & 99.89\% & 0.99 & 2,333   & 10.28     & 0.40   & 0.07                 \\ \hline  \vspace{2mm}\\[-5mm]\hline 
\multirow{7}{*}{ILSVRC2012} & PGD               & 97.70\%  & 0.91 & 168,919 & 5,451.51 & 14.92 & 0.06                 \\
                            & C\&W              & 99.33\%  & 0.95 & 200,256 & 299 & 1.13 & 0.11                 \\
                            & JSMA              & 90.00\%  & 0.94 & 447    & 75,375.33  & 194.35 & 1.00                 \\
                            & One-pixel         & 40.56\%  & 0.99 & 15     & 29.43     & 8.69   & 1.00                 \\
                            & Adv.Patch & 99.48\%  & 0.90 & 14,700  & 8,223.06  & 70.57 & 1.00                 \\
                            & LaVAN             & 95.10\%  & 0.96 & 7,500   & 5,985.29   & 73.53  & 1.00                 \\
                            & Ours              & 99.80\% & 0.99 & 98,431  & 168   & 1.33  & 0.08                 \\ 
\bottomrule 
\end{tabular}
\end{table*}

\subsection{Comparison among Adversarial Attacks}
To further evaluate the performance of our method, we report ASR, SSIM, and the  $\ell_{p}$ distortion of different attack methods in Table \ref{table1}. ResNet-18 and VGG-19 are leveraged to run on CIFAR-10 and ILSVRC2012 respectively. Six classic adversarial attack methods are compared with our method. For PGD, the perturbation bound $\epsilon=16/255$, the step size $\alpha=2/255$, and 20 iterations are adopted. For C\&W, the constant $c=1$, learning rate $lr=0.01$ and 1,000 iterations are adopted. Besides the global attacks PGD and C\&W, two local attacks JSMA and One-pixel attack are also considered. For JSMA, we set intensity variations $\theta=0.3$. For the One-pixel attack, we adopt five pixel-modification. 

It can be seen from Table \ref{table1} that our method outperforms other classic methods. We achieve 99.89\% ASR on CIFAR-10 and 99.80\% ASR on ILSVRC2012, which exceeds the state-of-the-art C\&W. We further analyze the distortion of adversarial examples with different methods. Among these methods, our method and One-pixel achieve the highest value of SSIM. It is not surprise that One-pixel performs well because its  constraint $\ell_{0}$ is only 15, which means only 5 pixels are modified. However, the number of modified pixels of our method is far more than that of One-pixel and JSMA. It seems contradict to the value of SSIM achieved by our method. We will further explain the reason through $\ell_{1}$,  $\ell_{2}$ and $\ell_{\infty}$ norms. As we known,  $\ell_{1}$ represents the sum of the absolute value of a perturbation on each pixel. In Table \ref{table1}, $\ell_{1}$ of ours is the smallest on CIFAR-10, and rank only second to One-pixel on ILSVRC2012, which means our total perturbations is the smallest. Similarly, $\ell_{2}$ and $\ell_{\infty}$ of ours is the smallest on CIFAR-10 and the second smallest on ILSVRC2012, where $\ell_{2}$ measure the move distance of an adversarial example from its original example and $\ell_{\infty}$ means the largest change in the total pixels. As a result, these metrics finely explain the reason for the largest SSIM achieved by our method. In conclusion, our method is powerful (higher ASR), as well as imperceptible (higher SSIM).

\subsection{Transferability}
\label{IV-E}
Transferability is an important property of adversarial examples, that is, the adversarial examples fooling one model can also fool other models \cite{[a22]}. This property is used to investigate the black-box attack ability of adversarial examples. In this section, we compare the transferability of our CFR patch examples with the other classic adversarial examples in black-box settings. In these methods, PGD, M-DI$^2$-FGSM, and C\&W are global adversarial attacks, while Adversarial Patch and LaVAN are local adversarial attacks. Note that the model for crafting adversarial examples is referred to as a \textit{substitute model}  (the first row of Table  \ref{table2} and  \ref{table3}) , while the model for testing adversarial examples is termed as a \textit{target model}  (the first column of Table  \ref{table2} and  \ref{table3}) in this paper. When the substitute model is consistent with the target model, i.e., the case in the diagonal of Table, it is equivalent to a white-box attack.  In this section, we focus on transferability and merely consider the case that the target model is different from the substitute model. 

\begin{table*}
\caption{Transferability (represented by the ASR on the target model) on CIFAR-10. The first row is the target models and the first column is the substitute models. The second column represents various attack methods: PGD ($s=20$, $\epsilon=16/255$, $\alpha=2/255$), M-DI$^2$-FGSM ($s=20$, $\epsilon=16/255$, $\alpha=2/255$, $p=0.5$), C\&W ($c=1$, $lr=0.01$, $iterations=1,000$), and Ours ($\tau=0.2$, $T=0.1$, $\eta=10$, $N=20$, $\beta=1$).}
\setlength{\tabcolsep}{6mm}   
\label{table2}
\centering
\begin{tabular}{ccccccc}
\toprule  
Model                      & Attack     & VGG-11           & VGG-13            & VGG-16           & ResNet-18         & ResNet-34         \\ \hline
\multirow{4}{*}{VGG-11}    & PGD        & 93.51\%          & 65.17\%           & 63.46\%          & 55.18\%           & 58.15\%           \\
                           & M-DI$^2$-FGSM & 99.10\%          & 90.69\%           & \textbf{90.11\%} & \textbf{87.79\%}  & 85.09\%           \\
                           & C\&W       & 93.94\%          & 45.23\%           & 46.46\%          & 39.70\%           & 49.25\%           \\
                           & Ours       & \textbf{99.92\%} & \textbf{91.00\%}  & 86.97\%          & 86.12\%           & \textbf{85.69\%}  \\ \hline
\multirow{4}{*}{VGG-13}    & PGD        & 61.33\%          & 95.81\%           & 75.08\%          & 61.36\%           & 58.91\%           \\
                           & M-DI$^2$-FGSM & 74.47\%          & 97.70\%           & 80.11\%          & 79.78\%           & 75.38\%           \\
                           & C\&W       & 62.13\%          & 95.72\%           & 64.11\%          & 61.95\%           & 61.98\%           \\
                           & Ours       & \textbf{79.78\%} & \textbf{98.80\%} & \textbf{90.39\%} & \textbf{80.38\%}  & \textbf{80.68\%}  \\ \hline
\multirow{4}{*}{VGG-16}    & PGD        & 56.57\%          & 71.72\%           & 90.43\%          & 58.59\%           & 56.78\%           \\
                           & M-DI$^2$-FGSM & 66.77\%          & 83.94\%           & 90.79\%          & 70.84\%           & 69.38\%           \\
                           & C\&W       & 51.52\%          & 52.53\%           & \textbf{96.97\%} & 40.40\%           & 48.48\%           \\
                           & Ours       & \textbf{81.88\%} & \textbf{88.29\%}  & 96.00\%          & \textbf{75.00\%}  & \textbf{77.08\%}  \\ \hline
\multirow{4}{*}{ResNet-18} & PGD        & 60.70\%          & 62.52\%           & 60.92\%          & 93.18\%           & 64.22\%           \\
                           & M-DI$^2$-FGSM & 72.83\%          & 82.12\%           & 83.42\%          & 94.60\%           & 84.35\%           \\
                           & C\&W       & 61.28\%          & 55.97\%           & 54.88\%          & 97.44\%           & 56.24\%           \\
                           & Ours       & \textbf{85.09\%} & \textbf{84.18\%}  & \textbf{85.03\%} & \textbf{99.89\%} & \textbf{89.59\%}  \\ \hline
\multirow{4}{*}{ResNet-34} & PGD        & 60.28\%          & 64.56\%           & 60.92\%          & 68.47\%           & 92.72\%           \\
                           & M-DI$^2$-FGSM & 78.08\%          & \textbf{87.78\%}  & 88.89\%          & 91.39\%           & 98.60\%           \\
                           & C\&W       & 57.75\%          & 47.40\%           & 46.00\%          & 48.59\%           & 90.34\%           \\
                           & Ours       & \textbf{84.95\%} & 86.29\%           & \textbf{89.00\%} & \textbf{90.95\%}  & \textbf{99.96\%} \\ 
\bottomrule 
\end{tabular}
\end{table*}

\begin{table*}
\caption{Transferability (represented by ASR on the target model) on ILSVRC2012. The first row is the target models and the first column is the substitute models. The second column represents various attack methods: PGD ($s=20$, $\epsilon=16/255$, $\alpha=2/255$), M-DI$^2$-FGSM ($s=20$, $\epsilon=16/255$, $\alpha=2/255$, $p=0.5$), C\&W ($c=1$, $lr=0.01$, $iterations=1,000$), C\&W ($c=1$, $lr=0.01$, $iterations=1,000$), and Ours ($\tau=0.2$, $T=0.1$, $\eta=20$, $N=20$, $\beta=1$).}
\setlength{\tabcolsep}{5.3mm}   
\label{table3}
\centering
\begin{tabular}{ccccccc}
\toprule  
Model                       & Attack            & VGG-16            & VGG-19            & ResNet-34         & ResNet-50        & ResNet-101        \\ \hline
\multirow{6}{*}{VGG-16}     & PGD               & 99.16\%           & 84.74\%           & 61.55\%           & 60.92\%          & 51.81\%           \\
                            & M-DI$^2$-FGSM        & 99.72\%           & 85.90\%           & 60.46\%           & 62.11\%          & 52.58\%           \\
                            & C\&W              & 95.81\%           & 80.44\%           & 61.55\%           & 61.45\%          & 59.56\%           \\
                            & Adv.Patch & 98.60\%           & 35.78\%           & 27.50\%           & 24.40\%          & 23.00\%           \\
                            & LaVAN             & 94.41\%           & 55.48\%           & 37.00\%           & 27.70\%          & 27.00\%           \\
                            & Ours              & \textbf{99.92\%} & \textbf{86.15\%}  & \textbf{67.12\%}  & \textbf{69.47\%} & \textbf{64.08\%}  \\ \hline
\multirow{6}{*}{VGG-19}     & PGD               & 86.73\%           & 97.70\%           & 61.14\%           & 57.63\%          & 54.52\%           \\
                            & M-DI$^2$-FGSM        & \textbf{88.41\%}  & 99.61\%           & 56.11\%           & 63.95\%          & 54.13\%           \\
                            & C\&W              & 77.37\%           & 99.33\%           & 60.46\%           & 60.26\%          & 58.40\%           \\
                            & Adv.Patch & 69.27\%           & 93.03\%           & 27.40\%           & 24.50\%          & 23.00\%           \\
                            & LaVAN             & 73.46\%           & 94.87\%           & 35.00\%           & 35.60\%          & 30.00\%           \\
                            & Ours              & 83.90\%           & \textbf{99.80\%} & \textbf{73.51\%}  & \textbf{67.89\%} & \textbf{66.67\%}  \\ \hline
\multirow{6}{*}{ResNet-34}  & PGD               & 75.70\%           & 75.87\%           & 99.18\%           & 68.68\%          & 58.27\%           \\
                            & M-DI$^2$-FGSM        & 70.39\%           & \textbf{76.20\%}  & \textbf{99.50\%} & 82.37\%          & 75.84\%           \\
                            & C\&W              & 74.16\%           & 74.90\%           & 92.53\%           & 61.71\%          & 58.40\%           \\
                            & Adv.Patch & 19.47\%           & 13.30\%           & 98.31\%           & 23.21\%          & 37.04\%           \\
                            & LaVAN             & 24.44\%           & 16.92\%           & 96.88\%           & 54.47\%          & 46.90\%           \\
                            & Ours              & \textbf{75.92\%}  & 74.86\%           & \textbf{99.50\%} & \textbf{82.84\%} & \textbf{83.33\%}  \\ \hline
\multirow{6}{*}{ResNet-50}  & PGD               & \textbf{75.70\%}  & 70.60\%           & 62.36\%           & 99.47\%          & 60.72\%           \\
                            & M-DI$^2$-FGSM        & 68.44\%           & 69.35\%           & \textbf{72.55\%}  & 99.34\%          & 84.37\%           \\
                            & C\&W              & 74.58\%           & \textbf{75.59\%}  & 61.82\%           & 94.08\%          & 57.75\%           \\
                            & Adv.Patch & 19.41\%           & 20.67\%           & 40.00\%           & 98.63\%          & 44.06\%           \\
                            & LaVAN             & 24.58\%           & 26.07\%           & 46.74\%           & 94.74\%          & 51.03\%           \\
                            & Ours              & 70.68\%           & 69.64\%           & 71.77\%           & \textbf{99.74\%} & \textbf{85.56\%}  \\ \hline
\multirow{6}{*}{ResNet-101} & PGD               & 74.30\%           & 68.65\%           & 60.19\%           & 68.03\%          & 99.48\%           \\
                            & M-DI$^2$-FGSM        & 61.87\%           & 57.42\%           & 69.84\%           & \textbf{85.39}\%          & 99.61\%           \\
                            & C\&W              & \textbf{77.51}\%           & 74.90\%           & 62.91\%           & 61.05\%          & 94.19\%           \\
                            & Adv.Patch & 20.26\%           & 28.30\%           & 44.05\%           & 60.09\%          & 98.13\%           \\
                            & LaVAN             & 26.20\%           & 31.31\%           & 52.31\%           & 68.82\%          & 94.19\%           \\
                            & Ours              & 76.87\%  & \textbf{76.06\%}  & \textbf{80.41\%}  & 81.13\% & \textbf{99.90\%} \\ 
\bottomrule 
\end{tabular}
\end{table*}

On CIFAR-10, ASR on the target models is shown in Table \ref{table2}, in which a higher value indicate higher transferability. For example, the adversarial examples generated by our method on the substitute model VGG-13,  have the highest ASR on other target models, which indicates our method has higher transferability than other attack methods, including state-of-the-art M-DI$^2$-FGSM. Besides VGG-13, on the other substitute models, the adversarial examples generated by our method have the highest transferability in most cases. On ILSVRC2012, the similar conclusion is obtained (as shown in Table \ref{table3}). In summary, our patch perturbation crafted with network explanations has a more powerful attack ability in black-box settings. We speculate the possible reason is that different classifiers share the similar CFR of an object, which is consistent with the result shown in  \cite{wu2020boosting}.

\subsection{ Adversarial Attacks on Protected Models}
The evaluation in Section \ref{IV-B}-\ref{IV-E} is conducted on the target model without protection. Now we further test the attack ability of our patch adversarial examples on protected models. We mainly focus on the model protected by adversarial training, because adversarial training is popularly considered as one of the most effective defenses \cite{shafahi2019adversarial}. Here we use Fast adversarial training \cite{[a16]} and PGD adversarial training \cite{[a4]}. For Fast adversarial training, we set the perturbation bound $\epsilon=8/255$, step size $\alpha=10/255$, and $epoch=20$ on CIFAR-10. Meanwhile, we set the perturbation bound $\epsilon=2/255$, step size $\alpha=2.5/255$, and $epoch=15$ on ILSVRC2012 respectively. For PGD adversarial training, we set 7 iteration steps, the step size $\alpha=2/255$, total perturbation bound $\epsilon=8/255$, and $epoch=15$. Finally, we obtain four protected models ResNet-18-Fast, ResNet-18-PGD, VGG-16-Fast, and VGG-16-PGD as target models.

For CIFAR-10, we compare our method with PGD and C\&W. For ILSVRC2012, we add two patch attacks Adv.Patch and LaVAN. Table \ref{table4} reports the results of the protected models under various attacks. We observe that adversarial training cannot achieve perfect performance against these attacks (ASR is higher than 50.00\%). Among them, our CFR patch attack outperforms other attacks in ASR. For example, our method can achieve 79.87\% ASR against VGG-16-Fast on ILSVRC2012. 

\begin{table}
\caption{ASR of various attack methods on the protected models}
\label{table4}
\centering
\begin{tabular}{cccc}
\toprule  

Dataset                                           & Protected Models                & Attack Methods    & ASR     \\ \hline
\multicolumn{1}{c}{\multirow{6}{*}{CIFAR-10}}    & \multirow{3}{*}{ResNet-18-Fast} & PGD               & 63.66\% \\
\multicolumn{1}{c}{}                             &                                 & C\&W              & 70.79\% \\
\multicolumn{1}{c}{}                             &                                 & Ours              & 79.00\% \\ \cline{2-4} 
\multicolumn{1}{c}{}                             & \multirow{3}{*}{ResNet-18-PGD}  & PGD               & 57.57\% \\
\multicolumn{1}{c}{}                             &                                 & C\&W              & 65.71\% \\
\multicolumn{1}{c}{}                             &                                 & Ours              & 77.39\% \\ \hline \vspace{2mm}\\[-5mm] \hline
\multicolumn{1}{c}{\multirow{10}{*}{ILSVRC2012}} & \multirow{5}{*}{VGG-16-Fast}    & PGD               & 65.99\% \\
\multicolumn{1}{c}{}                             &                                 & C\&W              & 66.11\% \\
\multicolumn{1}{c}{}                             &                                 & Adv.Patch & 63.89\% \\
\multicolumn{1}{c}{}                             &                                 & LaVAN             & 67.11\% \\
\multicolumn{1}{c}{}                             &                                 & Ours              & 79.87\% \\ \cline{2-4} 
\multicolumn{1}{c}{}                             & \multirow{5}{*}{VGG-16-PGD}     & PGD               & 63.50\% \\
\multicolumn{1}{c}{}                             &                                 & C\&W              & 57.72\% \\
\multicolumn{1}{c}{}                             &                                 & Adv.Patch & 75.75\% \\
\multicolumn{1}{c}{}                             &                                 & LaVAN             & 65.00\% \\
\multicolumn{1}{c}{}                             &                                 & Ours              & 77.60\% \\ 
\bottomrule 
\end{tabular}
\end{table}

\subsection{Analysis of Hyper-Parameters}
\begin{figure}
\centering
\subfigure[CIFAR-10]{
    \begin{minipage}[b]{0.35\textwidth}
    \includegraphics[width=1\textwidth]{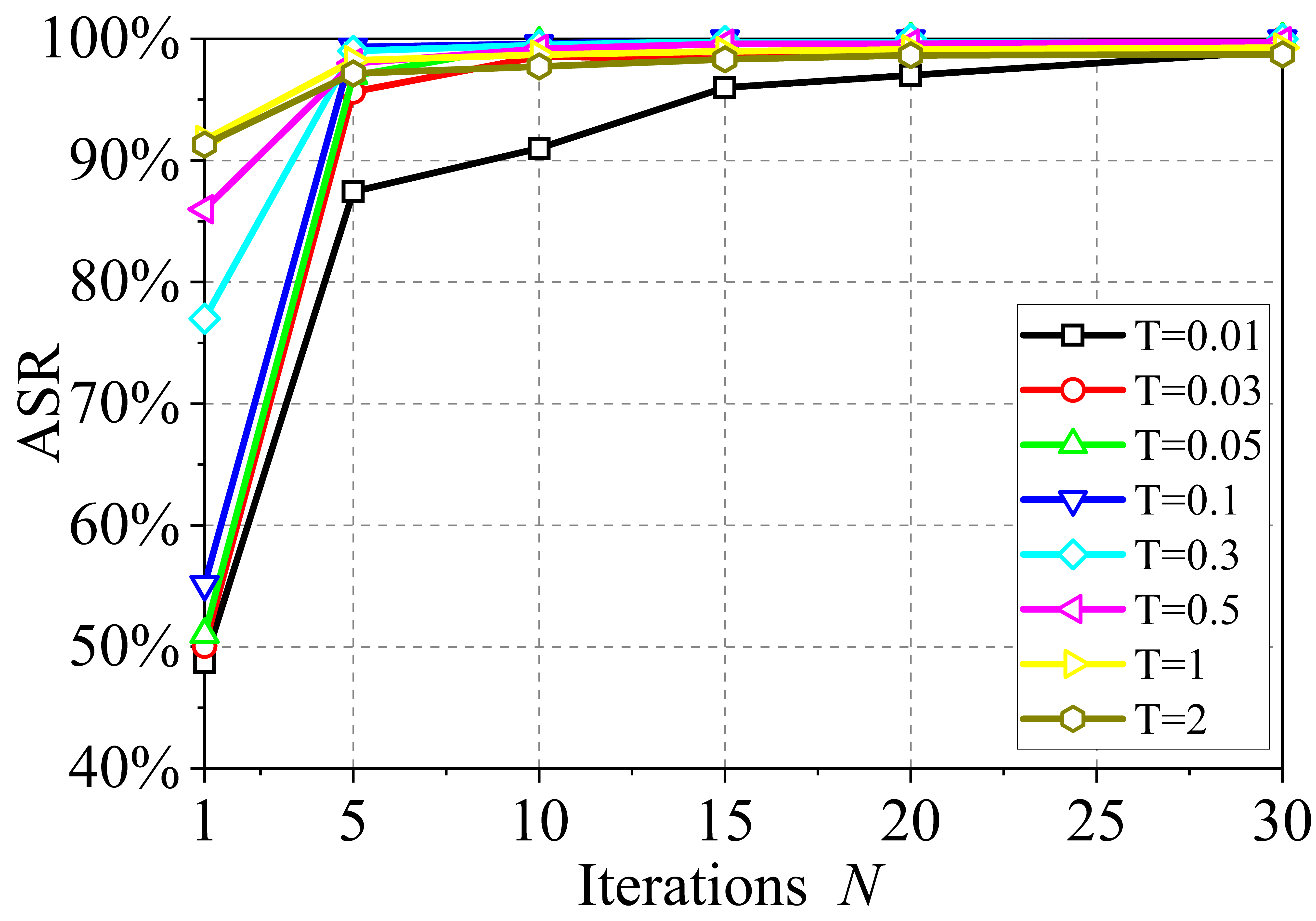}
    \end{minipage}
}
\subfigure[ILSVRC2012]{
    \begin{minipage}[b]{0.35\textwidth}
    \includegraphics[width=1\textwidth]{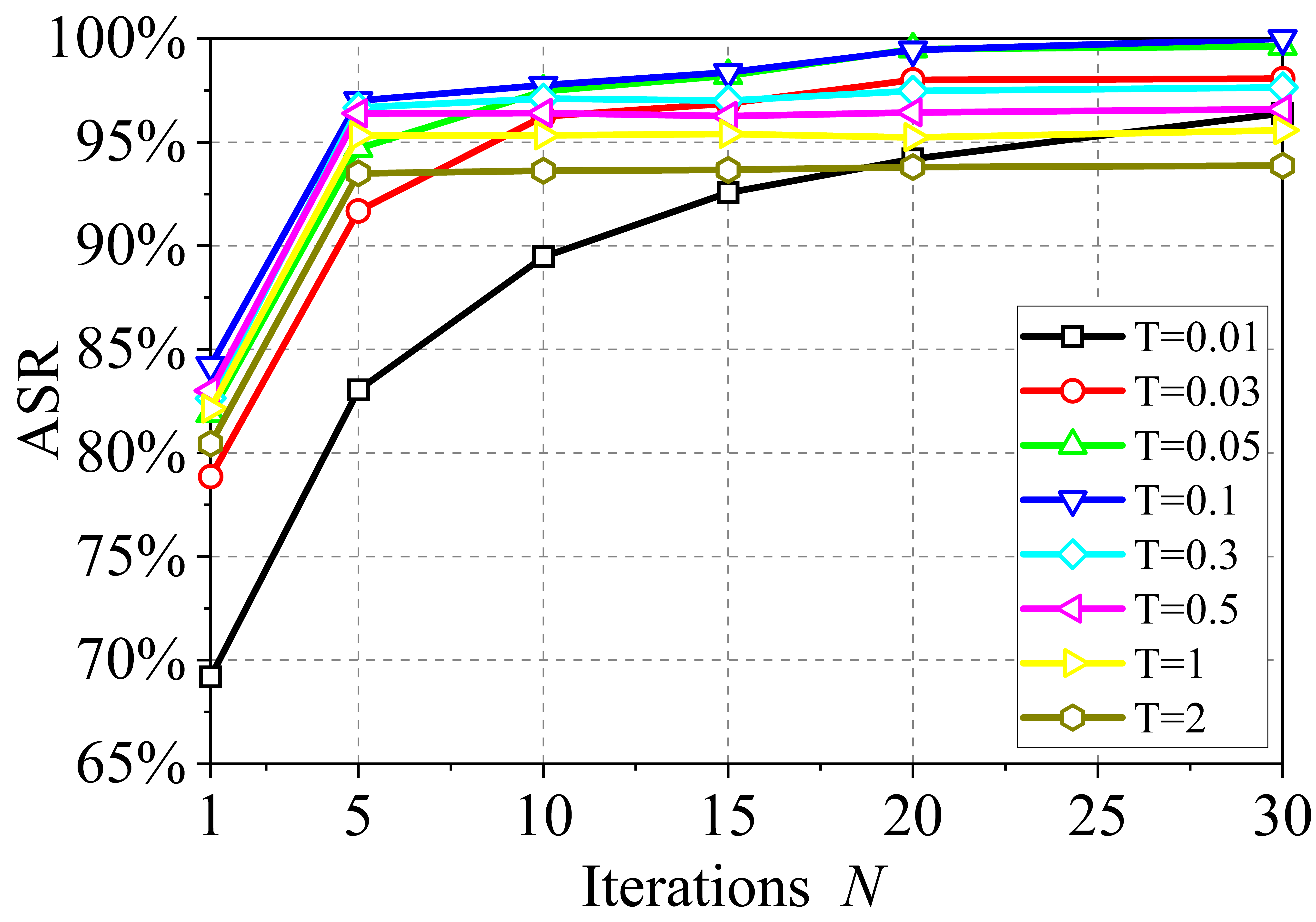}
    \end{minipage}
}
\caption{The effect of iterations $N$ and inverse temperature $T$ on the ASR. (a) ResNet-18 network on the CIFAR-10 ($\tau=0.2$, $\eta=10$, $\beta=1$); (b) VGG-16 network on the ILSVRC2012 ($\tau=0.2$, $\eta=20$, $\beta=1$).} 
\label{figa9}
\end{figure}

\textbf{Iterations $N$ and inverse temperature $T$} are two dominant hyper-parameters in our algorithm, and here we investigate their effects on ASR. We observe that ASR tends to increase along with iterations $N$ in Fig. \ref{figa9}. When $N$=30, ASR of our method can almost reach 100\% on both datasets with a proper inverse temperature (e.g. $T$=0.1), which indicates that our objective function can find the global optimal adversarial examples with fine-tuned parameters. We further discuss the impact of inverse temperature $T$. As shown in Fig. \ref{figa9}, when $T>1$ or $T$ is too small, it will prevent our patch attack from achieving a higher ASR regardless of increasing iterations. Remind that the purpose of inverse temperature $T$ is to prevent the loss $J_{CE}^{\prime}$ from decreasing to 0 as shown in Section \ref{sec3.4}. Nevertheless, when $T>1$, $J_{CE}^{\prime}$ becomes smaller, which leads to a smaller ASR, e.g., when $T=2$, it achieves the lowest ASR. Similarly, the smaller $T$ makes $J_{CE}^{\prime}$ become so large that it deviates far from the original value and the optimal direction, which makes it hard to converge to the optimal solution. For instance, when $T<0.1$, the ASR of patch attacks presents a downward trend. In summary, a moderate value of $T$ is desirable, e.g., $T=0.1$ for ILSVRC2012.

\begin{figure}
\centering
\subfigure[CIFAR-10]{
    \begin{minipage}[b]{0.35\textwidth}
    \includegraphics[width=1\textwidth]{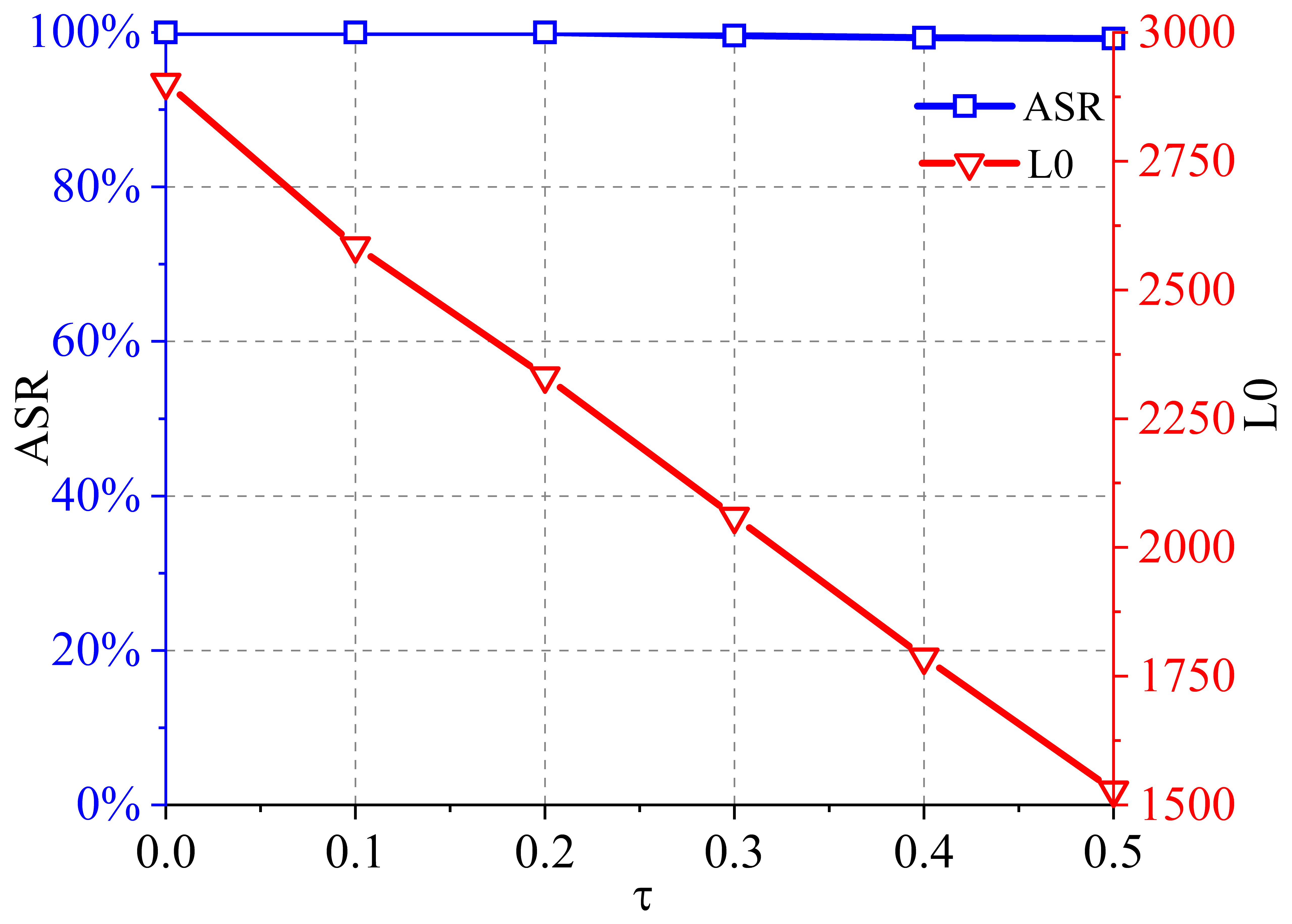}
    \end{minipage}
}
\subfigure[ILSVRC2012]{
    \begin{minipage}[b]{0.35\textwidth}
    \includegraphics[width=1\textwidth]{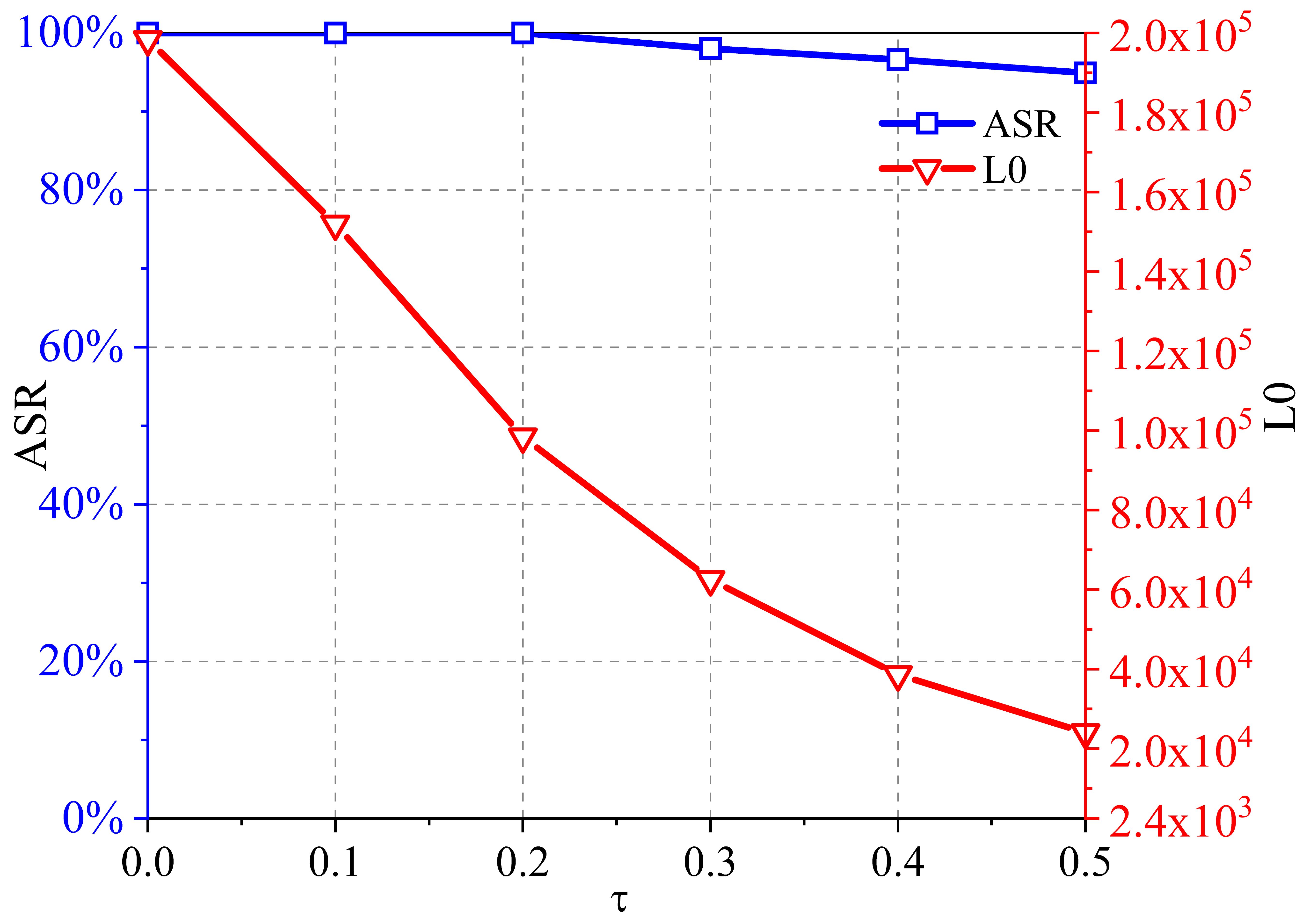}
    \end{minipage}
}
\caption{The influence of the threshold $\tau$ on the ASR and the $\ell_{0}$ norm of perturbations. (a) ResNet-18 network on CIFAR-10 ($N=30$, $T=0.1$, $\eta=10$, $\beta=1$); (b) VGG-16 network on ILSVRC2012 ($N=30$, $T=0.1$, $\eta=20$, $\beta=1$).} 
\label{figa10}
\end{figure}

\textbf{Threshold $\tau$} is also a dominant hyper-parameter to determines the shape of CFRs, that is, the size of the range of added perturbations. We use $\ell_{0}$ norm to measure the number of perturbed pixels. Specifically, $\tau=0$ means all pixels in the image are perturbed. As shown in Fig. \ref{figa10}, ASR and $\ell_{0}$ are presented in the same figure with two independent vertical axes. We observe that increasing the threshold $\tau$ can decrease $\ell_{0}$ norm, i.e., the size of perturbed regions is reduced, however, it does not affect ASR of our patch attacks. The reason is that the most contributing pixels are maintained all the time despite the size of perturbed regions changed with $\tau$. In other words, it reconfirms that the classification is mainly determined by the most contributing pixels.

\section{Conclusions}
\label{sec5}
Extensive experiments conducted on CIFAR-10 and ILSVRC2012 show that our patch attack outperforms existing global region attacks and local region attacks. Specifically, our patch attack has a higher attack success rate in both white-box and black-box settings. The main reason is that CFR plays a critical role in classification, and CFR of the same object is shared by multiple classification models. In addition, our crafted perturbations on CFR are imperceptible to human eyes. This imperceptible property is very attractive, which can be finely extended to other fields. For example, in the future we will utilize this imperceptibility to explore new adversarial patch attack on medical image classification or segmentation, which requires very tiny perturbations to evade medical specialists. Meanwhile, we are going to develop new countermeasures against the CFR patches, such as smoothing CFR to filter adversarial perturbations.

\section*{Acknowledgment}
This work is supported by National Key R\&D Program of China ( No.2018YFB2100400), Natural Science Foundation of China (No. 61972357), and Zhejiang Key R\&D Program (No. 2019C03135).

\ifCLASSOPTIONcaptionsoff
  \newpage
\fi



\bibliographystyle{IEEEtran}
\bibliography{IEEEexample}

\begin{thebibliography}{10}
\providecommand{\url}[1]{#1}
\csname url@samestyle\endcsname
\providecommand{\newblock}{\relax}
\providecommand{\bibinfo}[2]{#2}
\providecommand{\BIBentrySTDinterwordspacing}{\spaceskip=0pt\relax}
\providecommand{\BIBentryALTinterwordstretchfactor}{4}
\providecommand{\BIBentryALTinterwordspacing}{\spaceskip=\fontdimen2\font plus
\BIBentryALTinterwordstretchfactor\fontdimen3\font minus
  \fontdimen4\font\relax}
\providecommand{\BIBforeignlanguage}[2]{{%
\expandafter\ifx\csname l@#1\endcsname\relax
\typeout{** WARNING: IEEEtran.bst: No hyphenation pattern has been}%
\typeout{** loaded for the language `#1'. Using the pattern for}%
\typeout{** the default language instead.}%
\else
\language=\csname l@#1\endcsname
\fi
#2}}
\providecommand{\BIBdecl}{\relax}
\BIBdecl

\bibitem{[c10]}
A.~Krizhevsky, I.~Sutskever, and G.~E. Hinton, ``Imagenet classification with
  deep convolutional neural networks,'' \emph{Communications of the ACM},
  vol.~60, no.~6, pp. 84--90, 2017.

\bibitem{[c11]}
K.~Simonyan and A.~Zisserman, ``Very deep convolutional networks for
  large-scale image recognition,'' \emph{arXiv preprint arXiv:1409.1556}, 2014.

\bibitem{[c12]}
K.~He, X.~Zhang, S.~Ren, and J.~Sun, ``Deep residual learning for image
  recognition,'' in \emph{Proceedings of the IEEE Conference on Computer Vision
  and Pattern Recognition}, 2016, pp. 770--778.

\bibitem{[d1]}
Y.~Taigman, M.~Yang, M.~Ranzato, and L.~Wolf, ``Deepface: Closing the gap to
  human-level performance in face verification,'' in \emph{Proceedings of the
  IEEE Conference on Computer Vision and Pattern Recognition}, 2014, pp.
  1701--1708.

\bibitem{[c13]}
J.~Devlin, M.-W. Chang, K.~Lee, and K.~Toutanova, ``Bert: Pre-training of deep
  bidirectional transformers for language understanding,'' \emph{arXiv preprint
  arXiv:1810.04805}, 2018.

\bibitem{[d2]}
Y.~Goldberg, ``Neural network methods for natural language processing,''
  \emph{Synthesis Lectures on Human Language Technologies}, vol.~10, no.~1, pp.
  1--309, 2017.

\bibitem{[a1]}
C.~Szegedy, W.~Zaremba, I.~Sutskever, J.~Bruna, D.~Erhan, I.~Goodfellow, and
  R.~Fergus, ``Intriguing properties of neural networks,'' \emph{arXiv preprint
  arXiv:1312.6199}, 2013.

\bibitem{[m]}
I.~J. Goodfellow, J.~Shlens, and C.~Szegedy, ``Explaining and harnessing
  adversarial examples,'' \emph{arXiv preprint arXiv:1412.6572}, 2014.

\bibitem{[c15]}
T.~Tanay and L.~Griffin, ``A boundary tilting persepective on the phenomenon of
  adversarial examples,'' \emph{arXiv preprint arXiv:1608.07690}, 2016.

\bibitem{su2018robustness}
D.~Su \emph{et~al.}, ``Is robustness the cost of accuracy?--a comprehensive
  study on the robustness of 18 deep image classification models,'' in
  \emph{Proceedings of the European Conference on Computer Vision (ECCV)},
  2018, pp. 631--648.

\bibitem{[a3]}
A.~Kurakin, I.~Goodfellow, and S.~Bengio, ``Adversarial machine learning at
  scale,'' \emph{arXiv preprint arXiv:1611.01236}, 2016.

\bibitem{[a4]}
A.~Madry, A.~Makelov, L.~Schmidt, D.~Tsipras, and A.~Vladu, ``Towards deep
  learning models resistant to adversarial attacks,'' \emph{arXiv preprint
  arXiv:1706.06083}, 2017.

\bibitem{[h]}
N.~Carlini and D.~Wagner, ``Towards evaluating the robustness of neural
  networks,'' in \emph{Proceedings of the IEEE Symposium on Security and
  Privacy (SP)}, 2017, pp. 39--57.

\bibitem{[c3]}
N.~Papernot, P.~McDaniel, S.~Jha, M.~Fredrikson, Z.~B. Celik, and A.~Swami,
  ``The limitations of deep learning in adversarial settings,'' in \emph{2016
  IEEE European Symposium on Security and Privacy (EuroS\&P)}.\hskip 1em plus
  0.5em minus 0.4em\relax IEEE, 2016, pp. 372--387.

\bibitem{[c2]}
J.~Su, D.~V. Vargas, and K.~Sakurai, ``One pixel attack for fooling deep neural
  networks,'' \emph{IEEE Transactions on Evolutionary Computation}, vol.~23,
  no.~5, pp. 828--841, 2019.

\bibitem{[c24]}
T.~B. Brown, D.~Man{\'e} \emph{et~al.}, ``Adversarial patch,'' \emph{arXiv
  preprint arXiv:1712.09665}, 2017.

\bibitem{[c25]}
D.~Karmon, D.~Zoran, and Y.~Goldberg, ``Lavan: Localized and visible
  adversarial noise,'' \emph{arXiv preprint arXiv:1801.02608}, 2018.

\bibitem{[a12]}
R.~R. Selvaraju, M.~Cogswell, A.~Das, R.~Vedantam, D.~Parikh, and D.~Batra,
  ``Grad-cam: Visual explanations from deep networks via gradient-based
  localization,'' in \emph{Proceedings of the IEEE International Conference on
  Computer Vision}, 2017, pp. 618--626.

\bibitem{[a11]}
B.~Zhou, A.~Khosla, A.~Lapedriza, A.~Oliva, and A.~Torralba, ``Learning deep
  features for discriminative localization,'' in \emph{Proceedings of the IEEE
  Conference on Computer Vision and Pattern Rrecognition}, 2016, pp.
  2921--2929.

\bibitem{[a13]}
J.~Zhang, S.~A. Bargal, Z.~Lin, J.~Brandt, X.~Shen, and S.~Sclaroff, ``Top-down
  neural attention by excitation backprop,'' \emph{International Journal of
  Computer Vision}, vol. 126, no.~10, pp. 1084--1102, 2018.

\bibitem{[c14]}
S.~Zagoruyko and N.~Komodakis, ``Paying more attention to attention: Improving
  the performance of convolutional neural networks via attention transfer,''
  \emph{arXiv preprint arXiv:1612.03928}, 2016.

\bibitem{[k]}
T.~Deng and Z.~Zeng, ``Generate adversarial examples by spatially perturbing on
  the meaningful area,'' \emph{Pattern Recognition Letters}, vol. 125, pp.
  632--638, 2019.

\bibitem{[c23]}
I.~Evtimov \emph{et~al.}, ``Robust physical-world attacks on machine learning
  models,'' \emph{arXiv preprint arXiv:1707.08945}, 2017.

\bibitem{[d3]}
J.~He, Z.~Deng, and Y.~Qiao, ``Dynamic multi-scale filters for semantic
  segmentation,'' in \emph{Proceedings of the IEEE International Conference on
  Computer Vision}, 2019, pp. 3562--3572.

\bibitem{[c22]}
C.~Xie \emph{et~al.}, ``Adversarial examples for semantic segmentation and
  object detection,'' in \emph{Proceedings of the IEEE International Conference
  on Computer Vision}, 2017, pp. 1369--1378.

\bibitem{[c16]}
Z.~Gu \emph{et~al.}, ``Gradient shielding: Towards understanding vulnerability
  of deep neural networks,'' \emph{IEEE Transactions on Network Science and
  Engineering}, 2020.

\bibitem{wu2020boosting}
W.~Wu, Y.~Su, X.~Chen, S.~Zhao, I.~King, M.~R. Lyu, and Y.-W. Tai, ``Boosting
  the transferability of adversarial samples via attention,'' in
  \emph{Proceedings of the IEEE/CVF Conference on Computer Vision and Pattern
  Recognition}, 2020, pp. 1161--1170.

\bibitem{[c17]}
A.~Kurakin, I.~Goodfellow, and S.~Bengio, ``Adversarial examples in the
  physical world,'' \emph{arXiv preprint arXiv:1607.02533}, 2016.

\bibitem{[c18]}
F.~Tram{\`e}r, A.~Kurakin, N.~Papernot, I.~Goodfellow, D.~Boneh, and
  P.~McDaniel, ``Ensemble adversarial training: Attacks and defenses,''
  \emph{arXiv preprint arXiv:1705.07204}, 2017.

\bibitem{[c19]}
Y.~Dong, F.~Liao, T.~Pang, H.~Su, J.~Zhu, X.~Hu, and J.~Li, ``Boosting
  adversarial attacks with momentum,'' in \emph{Proceedings of the IEEE
  Conference on Computer Vision and Pattern Recognition}, 2018, pp. 9185--9193.

\bibitem{[c8]}
C.~Xie, Z.~Zhang, Y.~Zhou, S.~Bai, J.~Wang, Z.~Ren, and A.~L. Yuille,
  ``Improving transferability of adversarial examples with input diversity,''
  in \emph{Proceedings of the IEEE Conference on Computer Vision and Pattern
  Recognition}, 2019, pp. 2730--2739.

\bibitem{[c20]}
S.-M. Moosavi-Dezfooli, A.~Fawzi, and P.~Frossard, ``Deepfool: a simple and
  accurate method to fool deep neural networks,'' in \emph{Proceedings of the
  IEEE Conference on Computer Vision and Pattern Recognition}, 2016, pp.
  2574--2582.

\bibitem{[c1]}
N.~Akhtar and A.~Mian, ``Threat of adversarial attacks on deep learning in
  computer vision: A survey,'' \emph{IEEE Access}, vol.~6, pp.
  14\,410--14\,430, 2018.

\bibitem{[c4]}
K.~Xu, S.~Liu, P.~Zhao, P.-Y. Chen, H.~Zhang, Q.~Fan, D.~Erdogmus, Y.~Wang, and
  X.~Lin, ``Structured adversarial attack: Towards general implementation and
  better interpretability,'' \emph{arXiv preprint arXiv:1808.01664}, 2018.

\bibitem{[c26]}
X.~Yuan, P.~He, Q.~Zhu, and X.~Li, ``Adversarial examples: Attacks and defenses
  for deep learning,'' \emph{IEEE Transactions on Neural Networks and Learning
  Systems}, vol.~30, no.~9, pp. 2805--2824, 2019.

\bibitem{[c27]}
N.~Papernot, P.~McDaniel, I.~Goodfellow, S.~Jha, Z.~B. Celik, and A.~Swami,
  ``Practical black-box attacks against machine learning,'' in
  \emph{Proceedings of the 2017 ACM on Asia Conference on Computer and
  Communications Security}, 2017, pp. 506--519.

\bibitem{[c9]}
G.~Hinton, O.~Vinyals, and J.~Dean, ``Distilling the knowledge in a neural
  network,'' \emph{arXiv preprint arXiv:1503.02531}, 2015.

\bibitem{[a20]}
A.~Krizhevsky and G.~Hinton, ``Learning multiple layers of features from tiny
  images,'' \emph{Handbook of Systemic Autoimmune Diseases}, vol.~1, no.~4,
  2009.

\bibitem{[a17]}
O.~Russakovsky, J.~Deng, H.~Su, J.~Krause, S.~Satheesh, S.~Ma, Z.~Huang,
  A.~Karpathy, A.~Khosla, M.~Bernstein \emph{et~al.}, ``Imagenet large scale
  visual recognition challenge,'' \emph{International Journal of Computer
  Vision}, vol. 115, no.~3, pp. 211--252, 2015.

\bibitem{[d9]}
Z.~Wang, A.~C. Bovik, H.~R. Sheikh, and E.~P. Simoncelli, ``Image quality
  assessment: from error visibility to structural similarity,'' \emph{IEEE
  transactions on image processing}, vol.~13, no.~4, pp. 600--612, 2004.

\bibitem{[a22]}
N.~Papernot, P.~McDaniel, and I.~Goodfellow, ``Transferability in machine
  learning: from phenomena to black-box attacks using adversarial samples,''
  \emph{arXiv preprint arXiv:1605.07277}, 2016.

\bibitem{shafahi2019adversarial}
A.~Shafahi, M.~Najibi, A.~Ghiasi, Z.~Xu, J.~Dickerson, C.~Studer, L.~S. Davis,
  G.~Taylor, and T.~Goldstein, ``Adversarial training for free!'' \emph{arXiv
  preprint arXiv:1904.12843}, 2019.

\bibitem{[a16]}
E.~Wong, L.~Rice, and J.~Z. Kolter, ``Fast is better than free: Revisiting
  adversarial training,'' \emph{arXiv preprint arXiv:2001.03994}, 2020.

\end{thebibliography}
\end{document}